\documentclass[10pt,twocolumn,letterpaper]{article}

\usepackage[pagenumbers]{cvpr}
\usepackage{xspace}

\usepackage{microtype}

\makeatletter
\renewcommand\paragraph{\@startsection{paragraph}{4}{\z@}{0.6ex \@plus 0.3ex \@minus 0.2ex}{-1em}{\normalfont\normalsize\bfseries}}
\makeatother
\usepackage{array} %
\usepackage{multirow} %

\definecolor{cvprblue}{rgb}{0.21,0.49,0.74}
\usepackage[pagebackref,breaklinks,colorlinks,allcolors=cvprblue]{hyperref}

\newcommand{\method}{SOLVE\xspace}                   %

\title{SOLVE: Sparse, Ordered Keyposes Make LLM Agents \\ Versatile and Efficient Robot Manipulators}

\author
{
Xiang Li\textsuperscript{\rm 1,2*}\quad
Hongyu Zhou\textsuperscript{\rm 2*}\quad 
Ya-Li Li\textsuperscript{\rm 1}\quad 
Tao Li\textsuperscript{\rm 2}\quad 
Rui Zhang\textsuperscript{\rm 2}\quad \\
Junjie Lai\textsuperscript{\rm 2}\quad 
Weihua Zhang\textsuperscript{\rm 2$\ddagger$}\quad 
Shengjin Wang\textsuperscript{\rm 1$\ddagger$} \\
\textsuperscript{\rm 1}Department of Electronic Engineering, Tsinghua University \\
\textsuperscript{\rm 2}NVIDIA \\
}

\begin{document}
\twocolumn[{%
\renewcommand\twocolumn[1][]{#1}%
\maketitle
\vspace{-1.5em}
\begin{center}
  \includegraphics[width=\textwidth]{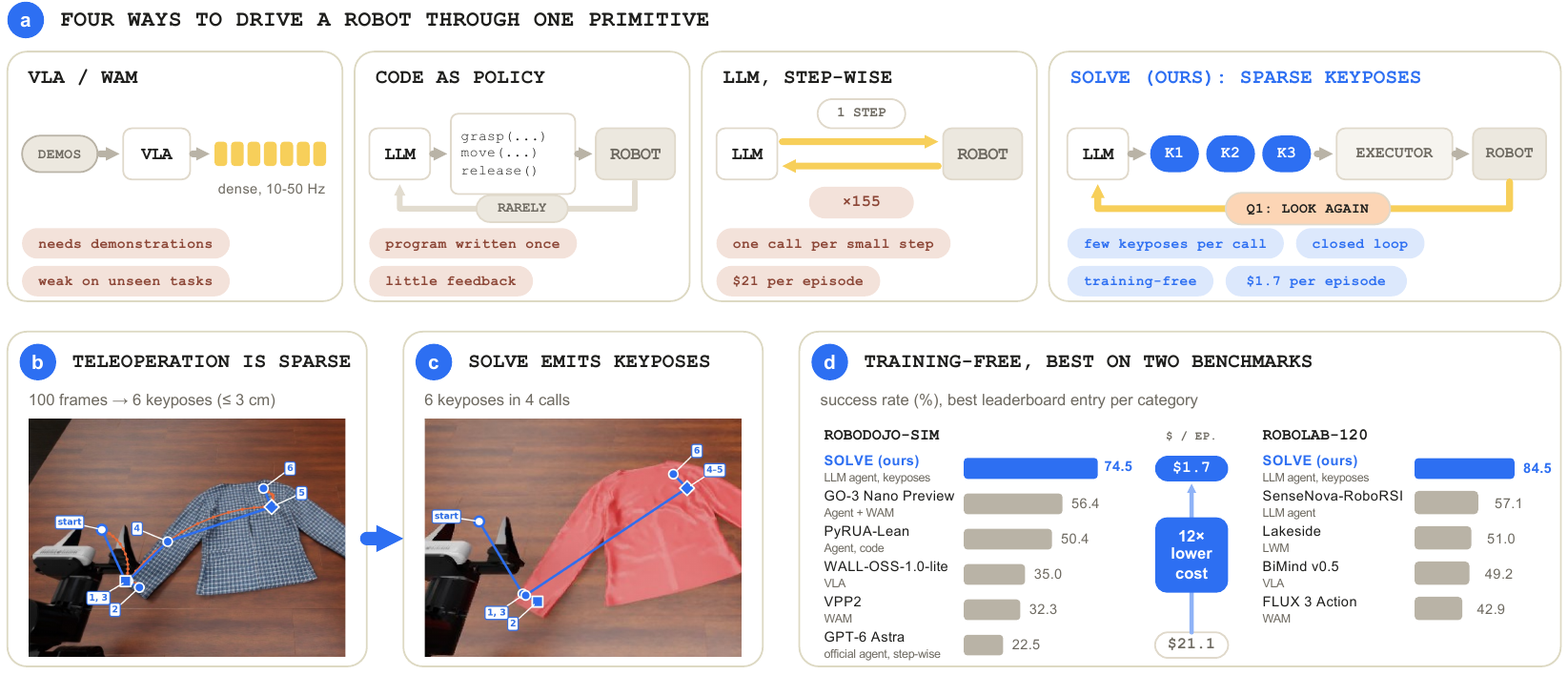}
  \captionof{figure}{%
    \textbf{Manipulation is sparse, and an LLM agent can exploit it.}
(a) VLAs/WAMs emit dense action chunks, code-as-policy runs a program with little feedback, and LLM agents spend one call per small step; SOLVE writes few keyposes per call. 
(b) A teleoperation compressed to 6 keyposes joined by straight lines. 
(c) SOLVE writes similar keyposes. 
(d) Success rate and cost against the best method.
    }
  \label{fig:teaser}
\end{center}
}]
\begingroup\renewcommand\thefootnote{}\footnotetext{\textsuperscript{*}Equal contribution.\quad \textsuperscript{$\ddagger$}Corresponding author.}\endgroup

\begin{abstract}

LLM agents have made remarkable progress in robot manipulation, yet they suffer from expensive inference, dense step-by-step control, and tool bottlenecks.
We introduce \method, a training-free, versatile and efficient agent policy: a multimodal LLM agent writes a sparse, ordered sequence of 6-DoF end-effector keyposes per call, observes the result, and writes the next, while a motion planner interpolates between them. The agent decides the number and spacing of keyposes for both fast long-range motion and fine-grained manipulation.
This granularity matches the sparsity of teleoperated manipulation: most demonstrations in DROID can be compressed into a few keyposes instead of dense actions.
\method grounds keyposes with sparse depth queries answered by noisy sources, such as multi-view triangulation, stereo, or monocular depth, and copes with that noise itself.
It also improves from execution feedback, reflecting on its own rollouts and distilling a few rules per task.
\method reaches state-of-the-art performance of 74.5\% on RoboDojo and 84.5\% on RoboLab, with only \$1.7 per episode on average, 12$\times$ less than previous dense-control agent policies.
On a real robot, the same agent achieves an 88\% success rate (75 of 85 episodes) over 16 challenging zero-shot tasks.
Our project page is at \href{https://serenos.github.io/solve-website/}{https://serenos.github.io/solve-website}.
\end{abstract}
    
\section{Introduction}
\label{sec:intro}

Robot manipulation policies have progressed from small task-specific networks trained with reinforcement or imitation learning~\cite{schulman2017proximal,haarnoja2018soft,zhao2023act,chi2023diffusionpolicy} to vision-language-action (VLA) models~\cite{brohan2023rt2,octo2024,kim2024openvla,black2025pi05} and world-action models (WAMs)~\cite{hu2024vpp,cen2026worldvla,vpp2}, yet all remain data-hungry: they need large teleoperation corpora for each robot and benchmark, and transfer poorly beyond them.
Frontier multimodal LLMs offer a training-free alternative and can now control robots directly~\cite{astranomy,gptaspolicy,astra,jia2026agp}, but current agents sit at one of two extremes.
Step-wise agents~\cite{astra,gptaspolicy,astranomy} issue one small step per model call, so every step costs seconds of latency and money, and errors compound.
Code-as-policy agents~\cite{liang2023cap,si2026pyrualean,gao2026hexashell} instead generate programs over a primitive library, which becomes a bottleneck: uncovered tasks require new primitives, written by hand or searched offline~\cite{gao2026hexashell,li2026embodiedevo}, and feedback arrives only when the program returns, too late to correct a failed step.

This raises a key question: \emph{what action granularity is appropriate for an agent policy?}
We compress 57,534 successful DROID episodes~\cite{khazatsky2024droid} into end-effector keyposes joined by straight lines (\cref{fig:teaser}b), and find that over 91\% of the travel lies on segments longer than 5\,cm, and that a grasp or release needs only 6 keyposes instead of 96 recorded actions.
We therefore hypothesize that, for an agent policy, a sequence of sparse end-effector keyposes suffices for manipulation, in place of a dense stream of actions.

We present \method, a training-free framework in which a multimodal LLM agent writes sparse, ordered end-effector keyposes, and a motion planner~\cite{sundaralingam2023curobo} executes straight-line motions between them (\cref{fig:teaser}c).
Each keypose is an absolute 6-DoF grasp-point pose with a gripper command; one call specifies up to five keyposes, roughly one primitive such as approach, grasp, and lift, and the gripper actuates only upon arrival.
The agent decides how many keyposes to use, how far apart, and how fast: a single keypose for a long free-space motion, and short, slow ones near contact.
After each sequence, the agent inspects the outcome before predicting the next one, so visual feedback enters its reasoning at every primitive rather than only at the checks a program anticipates.

Because each sparse keypose is an absolute commitment, it must be metrically accurate. \method therefore grounds keyposes through \emph{sparse metric queries}: the agent specifies the few pixels on which its next keypose depends, and a tool returns their 3D positions, each annotated with its measurement modality.
In RoboDojo and RoboLab simulation, which provide no stereo cameras, the tool triangulates each queried pixel across multiple views by epipolar matching, and falls back to monocular depth estimated by Depth Anything~3~\cite{lin2025da3}, scaled by the known table height, when no multi-view match is found; on the real robot, it directly uses depth from FoundationStereo~\cite{wen2025foundationstereo} on a stereo infrared camera.
Over multiple rounds of keypose reasoning, the agent adapts to the resulting depth noise by itself.
On difficult tasks, the agent further improves from its own execution feedback without training: it reflects on rollouts from layouts disjoint from evaluation, and consolidates the notes from several rounds into at most five bounded rules per task, which are provided to it at test time.

Since no component of \method is trained, and keyposes live in end-effector space while the planner resolves embodiment-specific execution, the same agent deploys on different robots without modification.
\method achieves 74.5\% on the dual-arm RoboDojo under the leaderboard's metric (\cref{tab:robodojo}) and 84.5\% on the single-arm Franka RoboLab~\cite{robolab}.
On a real UR12e, it completes 75 of 85 episodes across 16 challenging zero-shot tasks (\cref{tab:real}), including table cleaning, peg-in-hole, sketch drawing, and playing Gomoku against a human.
It does so at \$1.7 per episode on average, with one quarter of the model calls and 12$\times$ lower cost than a step-wise agent (\cref{fig:teaser}d).

Our contributions are as follows:
\begin{itemize}
  \item We show that teleoperated demonstrations compress into a small number of keyposes and, building on this finding, propose \method, a framework in which an LLM agent writes sparse, ordered keyposes for direct manipulation.
  \item We propose sparse metric queries, combining cross-view epipolar matching with depth estimation to ground keyposes accurately and efficiently in 3D from RGB in simulation, and use stereo depth on the real robot.
  \item We enable the agent to improve from rollouts, and show that \method transfers without modification across robots and benchmarks, achieving state-of-the-art results on RoboDojo, RoboLab, and a real robot at 12$\times$ lower cost.
\end{itemize}

\begin{figure*}[t]
  \centering
  \includegraphics[width=\textwidth]{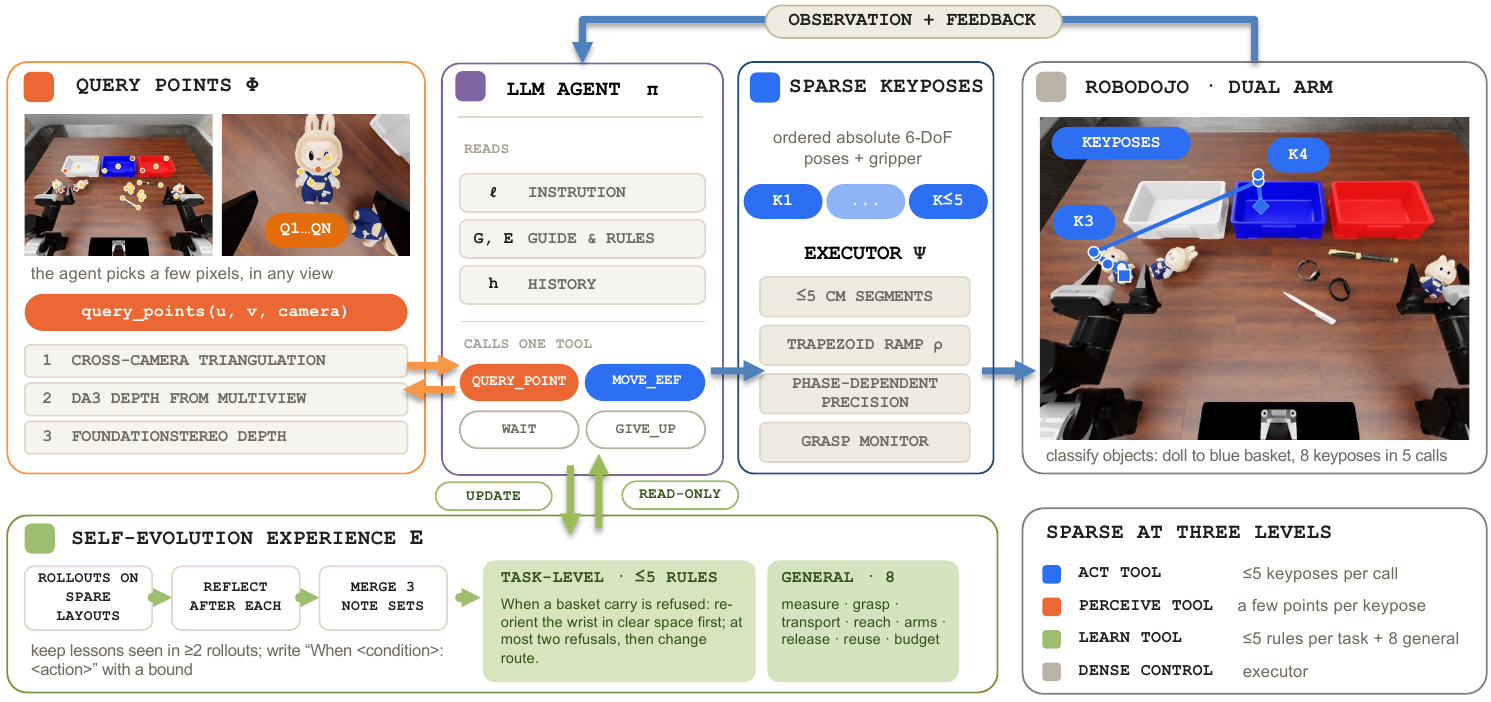}
  \caption{%
    \textbf{Overview of \method.}
    At each call, the multimodal LLM agent queries a few pixels, which return depth through epipolar matching across cameras or monocular depth estimation, and writes an ordered sequence of several absolute keyposes that an executor follows along straight lines.
    Offline, the agent reflects on its own rollouts, and the lessons are merged into general rules for self-improvement.}
    \vspace{-2pt}
  \label{fig:framework}
\end{figure*}

\section{Related Work}
\label{sec:related}

\paragraph{End-to-end policies.}
Vision-language-action (VLA) models~\cite{brohan2023rt1,brohan2023rt2,octo2024,kim2024openvla,black2024pi0,black2025pi05,liu2024rdt,kim2025openvlaoft,qu2025spatialvla,bu2025univla,shi2025hirobot,wen2025dexvla,zheng2026xvla,nvidia2025gr00tn1,team2025geminirobotics} and imitation policies that predict action chunks~\cite{zhao2023act,chi2023diffusionpolicy} are trained on large teleoperation corpora to emit dense, high-frequency actions; recent work improves their spatial representations~\cite{qu2025spatialvla,zheng2025tracevla,yang2025magma} and action tokenization~\cite{pertsch2025fast}.
World-action models (WAMs)~\cite{du2023unipi,wu2024gr1,cheang2024gr2,hu2024vpp,li2025uva,zhu2025uwm,jang2025dreamgen,cen2026worldvla,liao2026genieenvisioner,kim2026cosmospolicy,lyu2026lda1b,ye2026dreamzero,cai2026internvlaa1,vpp2} additionally learn from video to predict future observations and derive actions from them.
On RoboDojo~\cite{robodojo}, the best entries of both families reach 35\% on the leaderboard and at most 23\% on the Open dimension, whose tasks come without demonstrations.
\method instead acts through sparse keyposes and requires no robot-specific training.

\paragraph{Agentic policies.}
A growing line of work builds policies around frontier LLMs with little or no training.
\emph{Code-as-policy} methods let the model write programs over perception and control primitives~\cite{liang2023cap}; recent systems benchmark coding agents~\cite{fu2026capx}, structure programs as graphs rehearsed in simulation~\cite{chen2026gap}, grow skill libraries by evolutionary search~\cite{lu2026aspire}, or rewrite the training code of neural policies~\cite{xiao2026enpire}.
Agent as Policy~\cite{jia2026agp} similarly writes programs at test time on a real arm.
\emph{Harness} methods wrap a model with tools, memory and recovery logic, steering a frozen VLA as a retryable or interruptible tool~\cite{zhang2026harnessvla,chen2026volo,ding2026zetta} or calling semantic tools such as grasp and align~\cite{liu2026guava}.
Finally, some works use the model \emph{directly as a policy}, issuing one end-effector target per call~\cite{astra,gptaspolicy}, several absolute poses each executed for a single control step~\cite{astranomy}, discrete motion units of a few centimeters~\cite{chen2026showharness}, or batches of relative, axis-aligned motion commands.
\method is a direct policy in that every call observes the scene and outputs motion targets.
Unlike other direct policies, a call outputs an ordered sequence of absolute 6-DoF keyposes, placed with metric 3D measurements, that can span a whole primitive.

\paragraph{Sparse action representations.}
Sparsity itself is not new.
Keyframe policies predict the next gripper pose rather than a dense trajectory~\cite{james2022c2farm,shridhar2022peract,goyal2023rvt,gervet2023act3d,goyal2024rvt2,ke20243dda}, and bimanual policies condition on keyposes or skills~\cite{yu2024bikc,atomskill,gripperkeypose}.
These methods learn from task demonstrations, and they extract keyframes with heuristics such as gripper state changes and near-zero joint velocities~\cite{james2022c2farm,shridhar2022peract}, implicitly assuming that a few poses suffice; our analysis quantifies how many are needed per primitive and where this assumption fails.
VLM-based methods such as VoxPoser~\cite{huang2023voxposer}, ReKep~\cite{huang2024rekep}, and PIVOT~\cite{nasiriany2024pivot} are also training-free, but the model typically produces object-centric keypoints, constraints or value maps once, and a separate optimizer turns them into motion.
In \method, the model itself writes end-effector keyposes at every call (hence keypose rather than keypoint).

\section{Method}
\label{sec:method}

\subsection{Overview}
\label{sec:method:overview}
A robot with arms $\mathcal{A}$ is required to complete a task, given as a language instruction $\ell$, within $t_{\max}$ environment steps.
At step $t$ it observes $o_t$: RGB images from cameras $\mathcal{C}$ and, for each arm, the grasp-point pose $T^a_t\in SE(3)$ and gripper opening $g^a_t\in[0,1]$.
No depth is observed, but every camera pose follows from $o_t$ and a fixed calibration.
The agent is a frozen multimodal LLM $\pi$ (\cref{sec:exp_setup}) that emits exactly one tool call per model call $m$,
\begin{equation}
  z_m \sim \pi(\cdot \mid \ell,\, G,\, E,\, h_m), \qquad z_m \in \mathcal{Z},
  \label{eq:policy}
\end{equation}
given an optional task guide $G$ and experience notes $E$ (\cref{sec:method:experience}) and the history $h_m$ of earlier calls and results, in which only the latest two observations keep their images.
Of the tools $\mathcal{Z}$ (\cref{tab:tools}), \texttt{move\_eef} hands a keypose sequence $\mathcal{K}$ to the executor $\Psi$, which plays it and returns the next observation with execution feedback $e_m$; 
\texttt{query\_points} passes a few pixels $Q=\{(c_j,u_j)\}_j$, pixel $u_j$ of camera $c_j$, to the query tool $\Phi$, which returns their world coordinates without advancing the environment,
\begin{equation}
  \Phi(Q;\, o_t) = \{(X_j,\, \sigma_j)\}_j, \qquad X_j\in\mathbb{R}^3,
  \label{eq:query}
\end{equation}
where $X_j$ is the world point of pixel $j$ and $\sigma_j$ its source, triangulated, monocular or stereo depth, which tells the agent how far to trust it; \texttt{wait} lets the scene evolve, and \texttt{give\_up} ends the episode.

\subsection{Sparse, ordered keyposes}
\label{sec:method:keypose}

\paragraph{Why keyposes suffice.}
\label{sec:sparsity}

We first measure how far demonstrations compress into keyposes joined by straight lines.
Keeping the first and last poses and every gripper event (a grasp or a release), we simplify position (interpolated linearly) and orientation (by slerp) jointly with Ramer--Douglas--Peucker~\cite{douglas1973rdp}, adding the farthest frame as a keypose until every frame lies within a tolerance.
We apply it to RoboDojo's teleoperated demonstrations and to 57,534 successful DROID episodes~\cite{khazatsky2024droid}, which human operators teleoperated on real Franka arms.
\begin{figure}[t]
  \centering
  \includegraphics[width=\linewidth]{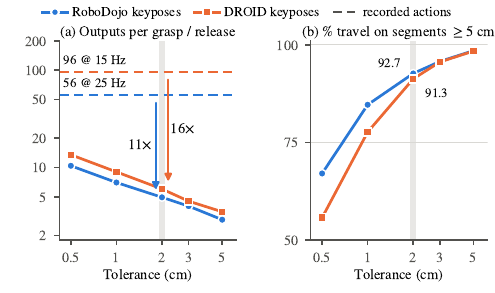}
  \caption{\textbf{Demonstrations compress into a few straight-line keyposes}, in simulation (RoboDojo) and on real robots (DROID).
  }
  \label{fig:compression}
\end{figure}
Both compress alike (\cref{fig:compression}).
A grasp or release spans a median of 56 recorded actions on RoboDojo (25\,Hz) and 96 on DROID (15\,Hz), as many as a policy imitating them at the recording rate must output; at 2\,cm / 10$^\circ$ it needs only 4.9 and 6.0 keyposes, 11$\times$ and 16$\times$ fewer, and a DROID episode shrinks to a median of 13 keyposes, about one per second.
Over 91\% of the travel lies on segments longer than 5\,cm, and on DROID this share stays within 88--95\% for every instruction verb.
The segments are genuinely straight: at 1\,cm, 92.6\% of DROID travel lies on segments whose path is at most 1.11 times their chord.
A short sequence of absolute targets thus describes most manipulation, and the dense motion can be left to the planner.

\begin{figure*}[t]
  \centering
  \includegraphics[width=\textwidth]{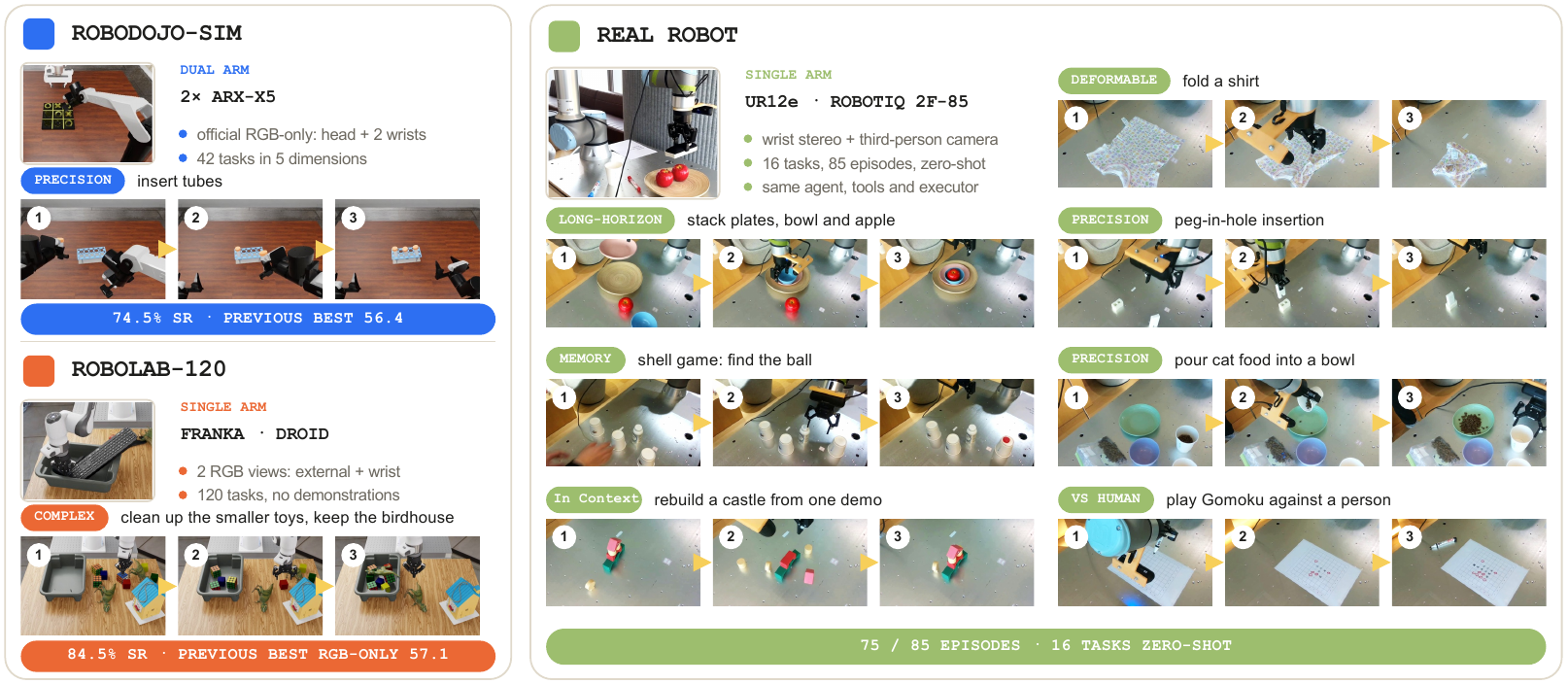}
  \caption{%
    \textbf{Three settings, three robots, one agent.}
    RoboDojo-Sim tests precision, long horizons and memory with two ARX-X5 arms, RoboLab tests unseen objects and instructions on a single Franka without demonstrations, and the real UR12e robot tests transfer to hardware on 16 unseen tasks.
    The same agent, tools and executor run in all three, with only the robot's constants changed.}
    \vspace{2pt}
  \label{fig:benchmarks}
\end{figure*}

\paragraph{Sparse, ordered keyposes.}
A keypose of arm $a\in\mathcal{A}$ is a target for its observed state $(T^a_t, g^a_t)$, $k^a = (\mathbf{p}^a, \boldsymbol{\theta}^a, g^a)$: the world-frame position $\mathbf{p}^a \in \mathbb{R}^3$ of the grasp point, its orientation $\boldsymbol{\theta}^a = (\text{pitch}, \text{roll}, \text{yaw})$, and the gripper opening $g^a \in [0, 1]$.
The agent writes only the dimensions it wants to change; all others hold their current value in $T^a_t$ and $g^a_t$, and naming dimensions of several arms moves them together.
One \texttt{move\_eef} call carries an ordered sequence $\mathcal{K} = (k_1, \dots, k_n)$ with $n \le n_{\max}$, a cap sized to one primitive, such as approach; descend and close; lift, which our demonstration analysis puts at five to six keyposes per grasp or release.
Three choices make the interface easy for an LLM to use.
\emph{Absolute targets:} each keypose is the pose the arm should reach, not an increment, so the agent never has to plan intermediate waypoints of a long move.
\emph{Arrive, then actuate:} a gripper command takes effect only after the arm reaches its keypose, so ``go there and grasp'' is a single keypose.
\emph{Agent-set speed:} each keypose may carry a ramp $r$, the number of environment steps used to accelerate toward it and to brake at it; the agent chooses a long ramp to carry liquid or set an object down gently, and a short one to strike or knock something over.
The agent thus decides the number, spacing and speed of keyposes: one keypose at full speed for a long free-space move, several short and slow ones near contact.
The executor $\Psi$ turns a keypose sequence $\mathcal{K}$ into joint commands without any learning. The move to each keypose is split into short segments of bounded translation and rotation, each solved for a joint path by a motion planner~\cite{sundaralingam2023curobo} and chained.
If a segment cannot be reached, the sequence stops before it and the feedback $e_m$ reports which keypose failed, so the agent can replan.

\subsection{Sparse metric queries}
\label{sec:method:queries}

Placing keyposes requires metric 3D grounding, but only at a few task-relevant points, not dense scene depth.
Through \texttt{query\_points}, the agent selects pixels $Q=\{(c_j,u_j)\}_j$ needed for its next keypose, such as a cup rim or a socket hole, and $\Phi$ returns their world points $X_j$ (\cref{eq:query}).
The tool uses calibrated RGB views in simulation with no stereo cameras, as described below, and additionally the stereo depth of the wrist camera on hardware (\cref{sec:exp_setup}).

\paragraph{Epipolar matching.}
We triangulate each query pixel by calibrated two-view matching along its epipolar line.
To accommodate the large viewpoint and scale differences between head and wrist cameras, we evaluate depth hypotheses using plane-induced patch warping and normalized cross-correlation rather than comparing fixed image windows.
This directly matches the agent-selected pixel without requiring it to be a detected keypoint or computing a dense disparity map.
We first try the other current cameras, then recent wrist frames with stricter acceptance thresholds to account for scene motion.
A point is accepted only when the match passes confidence, uniqueness and texture checks; otherwise, we use the depth model fallback below.
Implementation details and comparisons with feature matching are provided in the supplementary material.

\paragraph{Depth model fallback.}
When triangulation fails, \eg, on textureless surfaces or points visible in only one camera, the tool uses depth predicted by Depth Anything~3~\cite{lin2025da3} from all current views and their camera calibration.
We correct the scale per view using reference depths obtained by intersecting table-pixel rays with the known table plane.
The model runs once per observation $o_t$, shared by all queries; its depth is accurate to a centimeter but smooths edges and thin parts.

\paragraph{Telling the agent how much to trust.}
Each returned point $X_j$ carries its source $\sigma_j$ (\cref{eq:query}), triangulated, monocular or, on hardware, stereo; triangulated points additionally carry the match confidence.
The agent uses it to decide whether to act on a reading or to measure again, \eg, by querying a more distinctive pixel or the same point in a closer wrist view before a contact-critical keypose.

\begin{table*}[t]
\centering
\small
\setlength{\tabcolsep}{2pt}
\begin{tabular*}{\textwidth}{@{\extracolsep{\fill}}ll*{7}{c}@{}}
\toprule
 & & \multicolumn{2}{c}{Generalization} & & & & & \\
\cmidrule(lr){3-4}
Method & Type & Standard & Random & Precision & Long-Horizon & Memory & Open & \textbf{Average} \\
\midrule
GR00T N1.7~\cite{nvidia2025gr00tn1} & VLA & 2.4/4.0 & 0.0/0.3 & 0.7/2.5 & 3.6/8.3 & 0.9/1.1 & 0.2/0.2 & 1.3/2.9 \\
LingBot-VLA & VLA & 7.6/10.9 & 1.0/2.5 & 1.8/5.3 & 5.2/10.9 & 2.8/3.8 & 0.7/0.7 & 3.0/5.5 \\
$\pi_{0.5}$~\cite{black2025pi05} & VLA & 14.9/20.9 & 1.4/5.8 & 5.5/12.4 & 14.7/23.5 & 4.7/5.9 & 1.7/2.0 & 6.9/11.4 \\
WALL-OSS-1.0-lite & VLA & 45.4/53.4 & 30.4/39.9 & 29.1/38.2 & 32.2/49.9 & 56.0/57.4 & 19.6/20.8 & 35.0/42.6 \\
\midrule
InternW0-$\Delta$ & WAM & 33.8/41.0 & 11.8/19.2 & 23.2/32.0 & 29.3/46.3 & 34.0/34.7 & 10.2/10.8 & 23.9/30.8 \\
Awomo-0.5 & WAM & 30.9/37.7 & 16.8/22.9 & 33.5/39.6 & 26.8/40.4 & 41.2/42.3 & 22.9/24.1 & 29.6/35.3 \\
VPP2~\cite{vpp2} & WAM & 46.4/52.1 & 12.6/20.6 & 35.2/43.0 & 37.0/56.3 & 55.6/56.0 & 4.1/4.6 & 32.3/39.3 \\
\midrule
GPT-6 Astra (RoboProbe)~\cite{astra} & Agent & 32.7/35.3 & 28.3/31.4 & 4.0/12.7 & 8.2/21.4 & 38.7/43.0 & 31.0/34.4 & 22.5/29.0 \\
HexaShell~\cite{gao2026hexashell} & Agent & 38.0/43.3 & 31.0/35.1 & 32.0/41.2 & 31.5/42.9 & 81.0/84.5 & 55.2/58.0 & 46.9/53.2 \\
PyRUA-Lean~\cite{si2026pyrualean} & Agent & 45.3/50.5 & 41.0/46.1 & 38.2/44.8 & 43.8/54.3 & 68.3/68.4 & 58.2/60.6 & 50.4/55.3 \\
GO-3 Nano Preview & Agent+WAM & 43.0/45.9 & 24.3/27.0 & \underline{76.0/78.7} & \underline{53.5/68.2} & 63.7/64.5 & 55.2/61.7 & 56.4/61.9 \\
\midrule
\method$^\dagger$ & Agent & 56.7/60.6 & 50.0/53.2 & 60.0/65.6 & 40.0/56.6 & \textbf{96.7/96.7} & 57.5/59.6 & 61.5/67.1 \\
\method + experience$^\dagger$ & Agent & \underline{59.2/63.0} & \underline{55.9/59.0} & 72.5/75.3 & 46.2/61.5 & 91.2/92.4 & \underline{72.5/74.1} & \underline{68.0/72.9} \\
\method + experience, Astra$^\ddagger$ & Agent & \textbf{61.7/66.5} & \textbf{64.4/67.3} & \textbf{81.5/83.6} & \textbf{61.0/72.7} & \underline{94.3/94.8} & \textbf{72.8/76.5} & \textbf{74.5/78.9} \\
\bottomrule
\end{tabular*}
\caption{\textbf{RoboDojo-Sim.} Success rate / Score per dimension and their average. $^\dagger$Our two independent runs on seed~0,1, 5 episodes per task, with GPT-6.1 Sol. $^\ddagger$Official evaluation on seed~3, 2,100 episodes (50 per task). \textbf{Bold}: best, \underline{underline}: second best success rate.}

\vspace{0pt}
\label{tab:robodojo}
\end{table*}

\subsection{Self-evolution with experience}
\label{sec:method:experience}

The keypose interface, the executor and the queries are shared by all tasks and robots.
Task-specific knowledge enters only through the two optional inputs of \cref{eq:policy}, the guide $G$ and the experience notes $E$.

\paragraph{Task guides.}
Where a benchmark provides demonstrations, a guide $G$ explains one demonstration of the task in text and keyframes.
The LLM writes it offline from the demonstration's grasp-point trajectory and camera frames, following a fixed structure: the demonstration's stages and its hard parts, and then for each stage its goal, the moves as performed in \texttt{move\_eef} values, the grasp geometry, a check that tells when the stage is done, and head and wrist keyframes where the text cites them.
Every statement is tagged as general, such as which arm grasps which part of an object and in what order, or as specific to the demonstrated layout, such as its coordinates; 
A guide thus conveys a strategy in the agent's own action space rather than a trajectory to replay, since object poses differ in every evaluated layout.
Tasks without demonstrations, such as those of RoboLab and the real robot, run without a guide ($G=\varnothing$).

\paragraph{Self-evolution notes.}
Experience notes $E$ let the agent improve on a task from its own execution feedback, without parameter updates.
On layouts that do not overlap with the evaluation layouts, the agent runs $R$ rollouts of a task one after another.
After each one, with its history $h_m$ still in context, it is told whether it succeeded, shown the task's current notes, and answers with one more tool call $z_m$, \texttt{update\_experience}: a short summary of why the rollout succeeded or failed, and operations that add, update or delete notes.
The prompt asks only for lessons that decided the outcome of the task or of a sub-step, stated with the situation in which they hold rather than with coordinates, and requires any lesson that adds verification, retries or extra motion to bound it in attempts or steps, since the step budget $t_{\max}$ ends most failed episodes.
A note set $\mathcal{N}$ holds a bounded number of short notes, each with a few frames.

\paragraph{Merging into bounded rules.}
Raw notes vary from one writing run to the next, and their effect depends on what each run happens to write (\cref{sec:exp_ablation}); notes that urge the agent to verify or to be careful make it run out of steps.
We therefore write $S$ note sets $\mathcal{N}^{(1)},\dots,\mathcal{N}^{(S)}$ independently, each from its own $R$ rollouts, and merge them with one LLM call $\mathcal{M}$ per task,
\begin{equation}
  E_{\text{task}} = \mathcal{M}\big(\mathcal{N}^{(1)},\dots,\mathcal{N}^{(S)}\big), \qquad |E_{\text{task}}|\le L,
  \label{eq:experience}
\end{equation}
where $\mathcal{M}$ sees each set's final notes together with the outcome and summary of all $RS$ rollouts, and keeps only lessons that at least two rollouts support, citing them.
Each rule takes the form ``When \emph{situation}: \emph{action}'' under a word limit, with a step cost or limit wherever it spends steps; advice to verify without a bound and numbers specific to one layout are not allowed, and rules carry no frames.
A final call distills general rules $E_{\text{gen}}$ that recur across many tasks, on measurement, grasping, carrying, reachability, arm allocation, placing, merging moves and step budgeting.
At test time the agent reads $E=E_{\text{task}}\cup E_{\text{gen}}$.

\section{Experiments}
\label{sec:exp}

\subsection{Setup}
\label{sec:exp_setup}

All experiments share the agent code, tools and executor; only the robot's constants change (kinematic model, camera calibration, robot description in the prompt).
RoboDojo uses guides (except on Open) and self-evolution, RoboLab only self-evolution, and the real robot neither: each episode is a single zero-shot attempt.

\paragraph{RoboDojo.}
RoboDojo-Sim~\cite{robodojo} comprises 42 tasks across five dimensions on two ARX-X5 arms, with demonstrations except for Open tasks.
We use only RGB observations, follow the official evaluation protocol, and report success rate (SR) and partial-credit score.

\paragraph{RoboLab.}
RoboLab~\cite{robolab} comprises 120 tasks on a Franka; we use its external and wrist RGB views and follow the official protocol, without guides.

\paragraph{Real robot.}

We also run \method on a single-arm UR12e with a wrist stereo camera and a fixed third-person camera; \texttt{query\_points} additionally reads the wrist camera's stereo depth (details in the supplementary material).
We design 16 tasks (\cref{tab:real}) along the Long-Horizon, Precision, Memory and Open dimensions and beyond them: in-context imitation (rebuilding a structure that a person builds once in front of the robot), playing Gomoku against a person, and folding a shirt.
Several are hard: peg-in-hole insertion leaves only 1\,mm of clearance, and in drawing the agent picks up the pen itself, so the offset from the grasp point to the pen tip is unknown.
No task has pre-collected demonstrations, guides, notes or task-specific code.
We run five episodes per task and ten of shirt folding, 85 in all.

\begin{table}[t]
  \centering
  \small
  \setlength{\tabcolsep}{3pt}
  \begin{tabular}{@{}l p{0.58\linewidth}@{}}
    \toprule
    Tool & Function \\
    \midrule
    \texttt{move\_eef} & Up to five ordered keyposes (arm, position, orientation, gripper); executed by the planner (\cref{sec:method:keypose}). \\
    \texttt{query\_points} & Pixels of any camera image $\rightarrow$ world points, each tagged triangulated, monocular or stereo (\cref{sec:method:queries}). \\
    \texttt{wait} & Holds the arms for a number of steps, \eg, for an opponent's move. \\
    \texttt{give\_up} & Ends the episode as a failure. \\
    \midrule
    \texttt{update\_experience} & Offline only: adds, updates or deletes the task's notes after a rollout. \\
    \bottomrule
  \end{tabular}
  \caption{\textbf{Tools of \method.} Each model call emits exactly one tool.}
  \label{tab:tools}
\end{table}

\paragraph{Implementation details.}
\Cref{tab:tools} lists the tools. The prompt describes the robot, the keypose convention and the tools, then gives the instruction, $G$ and $E$; each later turn adds the current images, the arms' state and the last call's feedback.
One \texttt{move\_eef} call carries at most $n_{\max}=5$ keyposes (raised to 30 for drawing on the real robot), and the executor splits the move to each keypose into segments of at most 5\,cm and 20$^\circ$.
For self-evolution, each note set is written over $R=5$ rollouts and holds at most ten notes of at most 110 words and two frames each.
We write $S=3$ note sets per task, so the merge sees $RS=15$ rollouts, and keep at most $L=5$ rules of at most 70 words each; the final call yields eight general rules.

\subsection{Main results}
\label{sec:exp_main}

\begin{table}[t]
\centering
\small
\setlength{\tabcolsep}{1.2pt}
\begin{tabular*}{\linewidth}{@{\extracolsep{\fill}}ll*{4}{c}@{}}
\toprule
Method & Type & Simple & Mod. & Compl. & \textbf{All} \\
\midrule
$\pi_{0.5}$~\cite{black2025pi05} & VLA & 29.7 & 31.5 & 13.5 & 28.0 \\
HiDream-O1 & VLA & 43.1 & 37.7 & 32.9 & 39.9 \\
BiMind v0.5 & VLA & 51.6 & 52.6 & 32.4 & 49.2 \\
\midrule
Cosmos3-Nano & WAM & 40.6 & 35.4 & 25.3 & 36.8 \\
FLUX 3 Action & WAM & 49.1 & 39.2 & 28.2 & 42.9 \\
Lakeside & LWM & 57.6 & 44.1 & 42.4 & 51.0 \\
\midrule
VoLo$^\ddagger$ & Agent & 32.7 & 25.6 & 17.6 & 28.2 \\
Cosmic$^\ddagger$ & TAMP & 46.6 & 40.0 & 31.8 & 42.3 \\
SenseNova-RoboRSI~\cite{sensenova2026roborsi} & Agent & 59.4 & 56.2 & 50.6 & 57.1 \\
\midrule
\method & Agent & \underline{81.6} & \underline{79.0} & \underline{75.3} & \underline{79.8} \\
\method + experience & Agent & \textbf{85.0} & \textbf{84.6} & \textbf{82.4} & \textbf{84.5} \\
\bottomrule
\end{tabular*}
\caption{\textbf{RoboLab.} Success rate (\%) on the 120 tasks by difficulty tier. \method: five episodes per task. $^\ddagger$Observes depth.}

\vspace{0pt}
\label{tab:robolab}

\end{table}

\paragraph{RoboDojo.}
In the official evaluation (\cref{tab:robodojo}), \method with experience notes and GPT-6 Astra reaches 74.5\%, 18.1 points above the best entry, GO-3 Nano Preview (agent with a world-action model), and 24.1 above the strongest coding agent, PyRUA-Lean~\cite{si2026pyrualean}.
It leads in every dimension, while GPT-6 Astra (RoboProbe) drives the same model step by step to 22.5\%: the 52.0-point gap comes from the agent, not the model.
The margins are largest where learned policies must generalize: 61.7 and 64.4\% on standard and randomized Generalization, 15.3 and 23.4 points above the best entries, and 72.8\% on Open, which has no demonstrations and where no VLA or world-action model exceeds 22.9\%.
Its text history tracks what it has seen and done, giving 94.3\% on Memory.
Precision rises from 4.0\% for the step-wise agent to 81.5\%, ahead of GO-3 Nano Preview (76.0\%), as each keypose lands on a measured point.

With the cheaper GPT-6.1 Sol (mean of two runs, five episodes per task), \method reaches 61.5\% without notes and 68.0\% with them, both above every leaderboard entry.

\paragraph{Efficiency.}
Averaged over RoboDojo and RoboLab at list prices, \method costs \$0.5 per episode with GPT-6.1 Sol and \$1.7 with GPT-6 Astra, 42$\times$ and 12$\times$ less than the \$21.1 of GPT-6 Astra Direct~\cite{gptaspolicy}, a step-wise agent whose cost is public for 50 RoboDojo episodes.
On these, \method (Astra) succeeds in 74\% of them instead of 26\% with 36 instead of 154 model calls per episode.

\paragraph{RoboLab.}
On RoboLab (\cref{tab:robolab}), without guides or any change beyond the robot's constants, \method succeeds in 84.5\% of the episodes, 27.4 points above the best entry, SenseNova-RoboRSI~\cite{sensenova2026roborsi}, which uses GPT-6 Astra directly as the policy; its first rollout without notes already reaches 79.8\%.
The gap widens from 25.6 points on simple to 31.8 on complex tasks: \method loses only 2.6 points between them, the other methods 8.8--20.9.
Self-evolution improves 16 of the 30 tasks that the first rollout found hard (28 more successes); the rest fail on large or rounded objects that slip from the gripper, or on the step budget.

\begin{table}[t]
  \centering
  \small
  \setlength{\tabcolsep}{3pt}
  \begin{tabular*}{\linewidth}{@{\extracolsep{\fill}}lccrrr@{}}
    \toprule
    Task family & Tasks & Success & Calls & Keyposes & Min \\
    \midrule
    Long-Horizon      & 3 & 15/15 & 20.3 & 18.3 & 4.8 \\
    Precision         & 2 & 7/10 & 19.3 & 14.3 & 5.1 \\
    Memory            & 1 & 5/5 & 15.0 & 12.0 & 3.8 \\
    Open              & 5 & 22/25 & 24.0 & 119.6 & 7.9 \\
    In-context        & 2 & 10/10 & 25.0 & 40.0 & 8.7 \\
    Gomoku \vs human & 2 & 10/10 & 27.0 & 74.5 & 10.1 \\
    Shirt folding     & 1 & 6/10 & 17.0 & 26.0 & 3.5 \\
    \midrule
    All               & 16 & \textbf{75/85} & 21.9 & 57.3 & 6.6 \\
    \bottomrule
  \end{tabular*}
  \caption{\textbf{Real UR12e}, zero-shot. Successful episodes, and model calls, keyposes and minutes per episode.}
  \vspace{3pt}
  \label{tab:real}
\end{table}

\paragraph{Real robot.}

\method succeeds in 75 of 85 real-robot episodes (88\%) and completes each of the 16 tasks at least once (\cref{tab:real}), with 21.9 model calls and 6.6 minutes per episode.
It again sets the keypose density itself: three or four for a grasp and place, up to 30 per call when it draws.
In context, it rebuilds a six-cup pyramid and a five-part block castle after watching a person build each once; when the person has built the castle but not yet scattered the blocks, it keeps watching until they do. It also wins all ten Gomoku games against a person.
When an attempt falls short, it adjusts the next one from the images rather than repeating it: after a broken test stroke it lowers the pen by 1.5\,mm, and after a failed top-down grasp of a plate it tilts the gripper to grasp the rim.
No task-specific code prompts these corrections (reasoning quoted in the supplementary material).
Its ten failures come from a shirt sliding along with the gripper, an insertion released before both pegs are seated and then blocked by a protective stop, or a drawing that would drive the wrist camera into the table.

\subsection{Ablations}
\label{sec:exp_ablation}
\begin{table}[t]
\centering
\small
\setlength{\tabcolsep}{2.5pt}
\begin{tabular*}{\linewidth}{@{\extracolsep{\fill}}l*{6}{c}r@{}}
\toprule
Variant & G & P & LH & M & O & Avg & Calls \\
\midrule
\begin{tabular}[c]{@{}l@{}}GPT-6 Astra\\(RoboProbe)~\cite{astra}\end{tabular} & 32.7 & 4.0 & 8.2 & 38.7 & 31.0 & 22.9 & 154.6 \\
+ Sparse Keyposes & 40.0 & 32.5 & 37.5 & 92.5 & 45.0 & 49.5 & 31.9 \\
+ Query Point & 56.7 & 60.0 & 40.0 & \textbf{96.7} & 57.5 & 62.2 & 35.2 \\
+ Notes (\method) & \textbf{59.2} & \textbf{72.5} & \textbf{46.2} & 91.2 & \textbf{72.5} & \textbf{68.3} & 34.1 \\
\bottomrule
\end{tabular*}
\caption{\textbf{Ablations on RoboDojo}. SR (\%) per dimension; G is the standard Generalization split.
Calls: model calls per episode.
}

\vspace{3pt}
\label{tab:ablation}
\end{table}

\Cref{tab:ablation} adds the components of \method one at a time to GPT-6 Astra (RoboProbe)~\cite{astra}.

\paragraph{Sparse keyposes.}
Replacing its step-wise end-effector actions with sparse keypose sequences, run by our executor, while still estimating targets from the RGB images, raises the average success rate from 22.9 to 49.5\%.
It gains most on Memory (38.7 to 92.5\%), Long-Horizon (8.2 to 37.5\%) and Precision (4.0 to 32.5\%), where the step-by-step baseline runs out of steps before finishing.

\paragraph{3D queries.}
Measuring targets with \texttt{query\_points} raises it to 62.2\% at about the same cost, mostly where a centimeter matters: Precision rises from 32.5 to 60.0\% and standard Generalization from 40.0 to 56.7\%.
Offline, on 1,182 queries of the Precision tasks with simulator depth as reference, intersecting a pixel's ray with the table errs by 51.6\,mm (median) and object silhouettes by 20.0\,mm, while epipolar matching errs by 4.6\,mm on the 37\% of pixels it accepts; on holes, where silhouettes fail, it is about ten times more accurate (2.8 \vs 36.4\,mm on \emph{deposit coin}).
Online, triangulation and the Depth Anything~3 fallback raise the share of successful episodes from 55.5\% with silhouettes alone to 59.8\%, on par with simulator depth.

\paragraph{Experience notes.}
Merged notes raise the average to 68.3\% at no extra cost, mainly on Open (+15.0), Precision (+12.5) and Long-Horizon (+6.2).
Raw notes add only 3 points in two independent writing runs, and their gains depend on what each run writes; notes that urge more verification make the agent run out of steps or give up.
Merging three runs into at most five bounded rules per task removes such losses; on RoboLab, self-evolved notes likewise raise success from 79.8 to 84.5\% (\cref{tab:robolab}).

\subsection{Analysis}
\label{sec:exp_analysis}

\begin{figure}[t]
  \centering
  \includegraphics[width=0.7\linewidth]{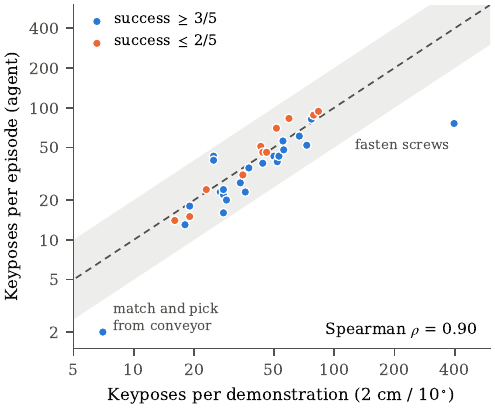}
  \caption{\textbf{Keyposes written by \method \vs keyposes a demonstration needs}, for the 34 RoboDojo tasks with demonstrations.
  }
  \label{fig:keyposes}
\end{figure}

\paragraph{Keyposes.}
\Cref{fig:keyposes} compares the keyposes \method writes per episode with those a teleoperated demonstration of the task needs.
The two counts agree closely: medians of 39.5 \vs 40.3 per task, a rank correlation of 0.90 over 34 tasks.
This is not merely because both grow with the number of objects: with the demonstration's grasps and releases held fixed, the partial rank correlation remains 0.69.

\begin{figure}[t]
  \centering
  \includegraphics[width=0.78\linewidth]{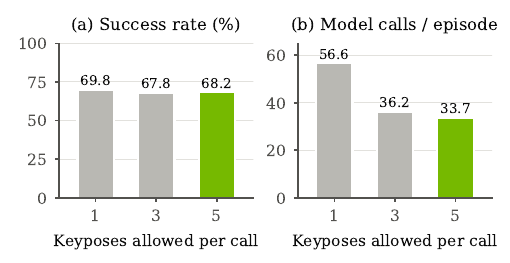}
  \caption{\textbf{Keyposes per call.} (a) Success rate and (b) model calls per episode against the keyposes allowed per \texttt{move\_eef} call.}
  \label{fig:keypose_cap}
\end{figure}

\paragraph{Keyposes per call.}
We cap a \texttt{move\_eef} call at one, three or five keyposes and change nothing else (\cref{fig:keypose_cap}).
Success is unchanged (69.8, 67.8 and 68.2\%), but model calls per episode fall from 56.6 to 36.2 and 33.7, 40\% fewer.
A higher cap would save little more in simulation: even with five allowed, the agent sends one to three keyposes in 89\% of its calls and five in only 3\%, as it often splits a primitive across calls to look in between.
Drawing needs many more: on the real robot, a cap of 30 lets the agent draw a portrait's 305 keyposes in 15 \texttt{move\_eef} calls instead of at least 61, each saving seconds of model latency.

\begin{figure}[t]
  \centering
  \includegraphics[width=0.68\linewidth]{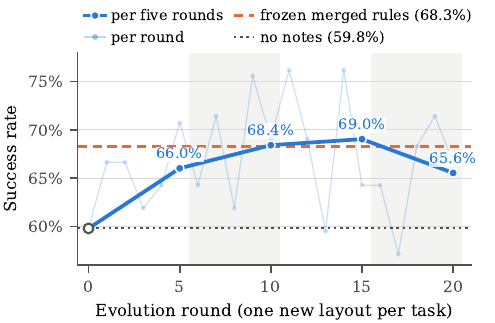}
  \caption{\textbf{Evolving the notes over more rounds.} Bold: SR per five rounds; dashed and dotted: our merged rules and no notes.}
  \label{fig:evolve}
\end{figure}

\paragraph{Evolving over more rounds.}
We let the agent keep evolving its notes for 20 rounds, each running one new layout per RoboDojo task (\cref{fig:evolve}).
Starting from 59.8\% without notes, success reaches 66.0, 68.4, 69.0 and 65.6\% after 5, 10, 15 and 20 rounds: it peaks near our frozen merged rules (68.3\%) and then declines.
It is also unstable: single rounds swing between 57.1\% and 76.2\%, while three runs with the frozen merged rules average 68.6\%.
The notes explain both: as they fill up, the agent stops adding and rewrites them around the latest episode, so that most notes end up describing one particular episode and earlier lessons are overwritten.
Merging avoids this: it keeps only layout-free lessons that recur across independent runs and freezes them, reaching the peak from the start and holding it across runs.

\section{Conclusion}
\label{sec:conclusion}

An LLM agent does not need dense, step-by-step control to manipulate. Teleoperated demonstrations compress to a few keyposes per grasp or release, and \method lets a multimodal LLM agent write such keyposes directly, grounds them with sparse metric queries, and refines them with notes from its own rollouts, without training.
\method reaches 74.5\% on RoboDojo and 84.5\% on RoboLab, ahead of every leaderboard entry, at \$1.7 per episode, and 88\% over 16 zero-shot real-world tasks.
Sparse, ordered keyposes are an effective and economical interface between LLMs and robot control.

\paragraph{Limitations.}
Some real tasks require force or continuous feedback that keyposes cannot provide, and each model call takes seconds; a next step is to combine keyposes with a learned low-level policy for contact-rich segments.

\clearpage

\section*{LLM Usage Statement}
Codex and Claude Code were used to help debug code and polish text and figures in the manuscript. 
All AI-assisted text, code, and experimental artifacts have been reviewed and verified by the authors, who take responsibility for the final content.

{
    \small
    \bibliographystyle{ieeenat_fullname}
    \bibliography{main}
}

\clearpage
\setcounter{page}{1}
\maketitlesupplementary
\appendix

\section{Overview}
\label{sec:suppl_overview}

This supplementary material complements the main paper in five parts.
\begin{itemize}\setlength\itemsep{1pt}
  \item \textbf{Implementation details} (\cref{sec:suppl_impl}): the settings of the agent loop, the executor, the query tool and the experience notes.
  \item \textbf{Additional analyses} (\cref{sec:suppl_analysis}): how far demonstrations compress into keyposes, how accurately query pixels are measured by epipolar matching and its fallbacks, what happens when notes keep evolving over many more rounds, and why the remaining RoboDojo episodes fail.
  \item \textbf{Prompts} (\cref{sec:suppl_prompts}): the system prompt, the tool descriptions, the prompts that write, merge and generalize experience notes, and the prompt that writes task guides.
  \item \textbf{More experiments} (\cref{sec:suppl_more}): controlled comparisons on a public RoboDojo subset (task inputs, reasoning effort and another model), the real-robot experiment in detail, and RoboTwin~2.0 as a third simulated benchmark.
  \item \textbf{Visualizations} (\cref{sec:suppl_vis}): keyframes of one episode of every real-robot task and of ten solved tasks of each simulated benchmark.
\end{itemize}
The supplementary archive also contains an offline webpage (open \texttt{index.html} in a browser) with videos of the real-robot episodes and of RoboDojo, RoboLab and RoboTwin rollouts.

\section{Implementation details}
\label{sec:suppl_impl}

\Cref{tab:impl} lists the settings of \method; they are the same for every task and, except where noted, every benchmark.

\begin{table}[t]
  \centering
  \footnotesize
  \setlength{\tabcolsep}{3pt}
  \begin{tabular*}{\linewidth}{@{\extracolsep{\fill}}l>{\raggedright\arraybackslash}p{0.58\linewidth}@{}}
    \toprule
    Setting & Value \\
    \midrule
    \multicolumn{2}{@{}l}{\emph{Agent}} \\
    Model, effort & GPT-6.1 Sol / GPT-6 Astra, \texttt{xhigh} \\
    Model calls per episode & $\le$600 ($\le$150 on RoboTwin) \\
    Tool calls per model call & exactly one \\
    Images kept & latest 2 observations \\
    \midrule
    \multicolumn{2}{@{}l}{\emph{Executor}} \\
    Keyposes per call $n_{\max}$ & 5 (30 for drawing on the real robot) \\
    Planned segment & $\le$5\,cm and $\le$20$^\circ$ \\
    Joint speed & $\le$0.1\,rad per step (25\,Hz) \\
    Ramp $r$ & 2 steps by default, 0--12 \\
    Settling tolerance & 0.002 / 0.01\,rad (contact / free) \\
    Slip detection & held $\ge$3\,mm, then closed \\
    \midrule
    \multicolumn{2}{@{}l}{\emph{Query tool}} \\
    Pixels per call & $\le$32 \\
    Depth hypotheses & $-$0.5 to 36\,cm above table, step 1.5\,mm \\
    Matching window & 15$\times$15, weighted NCC \\
    Second views & other cameras, 6 past wrist frames \\
    Depth estimation scale & median over $\ge$2,000 table pixels \\
    \midrule
    \multicolumn{2}{@{}l}{\emph{Experience}} \\
    Raw notes per set $\mathcal{N}$ & $\le$10, $\le$110 words, $\le$2 frames \\
    Writing & $S=3$ sets $\times$ $R=5$ rollouts \\
    Task rules $E_{\text{task}}$ & $L=5$ at most, $\le$70 words, $\ge$2 rollouts \\
    General rules $E_{\text{gen}}$ & 8, $\le$60 words, $\ge$5 tasks \\
    \bottomrule
  \end{tabular*}
  \caption{\textbf{Settings of \method}, shared by all tasks.}
  \label{tab:impl}
\end{table}

\paragraph{Agent loop.}
The agent is called through an OpenAI-compatible Responses API and must answer every turn with exactly one tool call; if a call has invalid arguments, such as a sequence of more than five keyposes, it is rejected with an explanation and the agent is prompted again.
Each call carries a \texttt{note}, one or two sentences on what the agent sees and why it acts, which is logged and kept in the history (\cref{sec:suppl_reasoning} quotes such notes).
Only the latest two observations keep their images (\cref{sec:method:overview}); older images are replaced by placeholders.

\paragraph{Feedback.}
After a \texttt{move\_eef} call, the next observation contains the current camera images, each arm's grasp-point pose and gripper opening, and the feedback $e_m$: for each arm, the distance from each commanded keypose; any slip (which stops that arm and skips its remaining keyposes) or fully closed grasp (which is reported without stopping); the keypose at which the planner refused a sequence, if any; and the remaining environment steps.
A move that would take more than 10\,s is refused and has to be split.
\texttt{wait} returns camera keyframes recorded while it held the arms, and can end early once the scene has stopped moving.

\paragraph{Writing and evaluation layouts.}
On both RoboDojo and RoboLab, experience notes are written only on layouts that differ from the evaluation layouts, then merged and frozen before evaluation, so no evaluation layout is seen while writing notes.
The \emph{first rollout} on RoboLab (\cref{sec:exp_main}) is the evaluation without notes; the self-evolved notes are written on other layouts and then used to evaluate the same evaluation layouts again.

\paragraph{Evaluation protocols of the main tables.}
The rows of \cref{tab:robodojo} follow two protocols.
The row with GPT-6 Astra is the official evaluation run by the RoboDojo team on seed~3, a hidden seed chosen by the organizers, with 50 episodes per task and 2,100 in all; the rows with GPT-6.1 Sol are the mean of two independent runs on seeds~0 and~1, five episodes per task.
As on the leaderboard, Average first averages Generalization over its standard and randomized splits, then averages the result with the other four dimensions.
\Cref{tab:ablation} keeps only the standard Generalization split and averages its five columns, so the same configuration has slightly different averages in the two tables: 22.5 \vs 22.9\% for GPT-6 Astra (RoboProbe), 61.5 \vs 62.2\% for \method without notes, and 68.0 \vs 68.3\% with the frozen merged rules.
The single-factor analyses, namely the online comparison of depth sources (\cref{sec:exp_ablation}), \cref{fig:keypose_cap} and the evaluations of \cref{fig:evolve}, use single runs on seed~0 and report the share of its 209 standard episodes that succeed: five layouts for each of the 42 tasks, less one layout of \emph{imitate sorting sequence} that the simulator marks unstable and never starts (\cref{sec:suppl_subset}).
\Cref{tab:robolab} runs five episodes for each of the 120 tasks, 600 in all.

\paragraph{Compute and cost.}
Each episode runs on one GPU, which serves the motion planner and Depth Anything~3; the model is called through its API.
Cost is computed from the logged token usage at the model's list price, including prompt-cache discounts.

\section{Additional analyses}
\label{sec:suppl_analysis}

\subsection{Keypose compression of demonstrations}
\label{sec:suppl_compression}

This section details the analysis of \cref{sec:sparsity}.

\paragraph{Data.}
RoboDojo provides 100 teleoperated demonstrations for each of its 34 tasks with demonstrations, recorded at 25\,Hz on two ARX-X5 arms; we convert each arm to the grasp point used by \texttt{move\_eef}, count the keyposes of both arms together, and skip an arm that moves less than 2\,cm and 5$^\circ$.
For DROID~\cite{khazatsky2024droid} we take all 57,639 successful episodes (15\,Hz, Franka with a Robotiq 2F-85) and drop those that move less than 2\,cm and 5$^\circ$ or last under one second, leaving 57,534 episodes and 18.7M frames, without subsampling.
We use DROID's recorded end-effector pose, detect gripper events with hysteresis (closed above 0.4, open below 0.2, ignoring flips shorter than three frames), and smooth positions over about 0.2\,s before measuring path lengths, which changes the shares of long segments by less than 0.3 points.

\paragraph{Robustness.}
On RoboDojo, every task has at least 80\% of its travel on straight segments longer than 5\,cm, the lowest being \emph{push T} (80.2\%) and \emph{pour liquid into cup} (80.4\%).
The 13,144 failed DROID episodes compress similarly: 88.9\% of their travel lies on long segments, against 91.3\% for successes.
Moving DROID's recorded pose 15\,cm along the tool axis, in case it is the flange rather than the fingertips, raises the median keyposes per episode from 13 to 15 and leaves the share at 91.2\%.
The analysis measures geometric compressibility only: neither the DROID nor the RoboDojo demonstrations record the states of the objects, so the compressed trajectories cannot be replayed to measure their success.

\subsection{Epipolar matching of query pixels}
\label{sec:suppl_sweep}

\texttt{query\_points} must return the 3D point of the pixel the agent chose, often the center of a hole or a point on a rim, from cameras that differ greatly in direction and scale.
We compare our epipolar matching (\cref{sec:method:queries}) with the alternatives offline, on pixels that the agent actually queried, using as reference the 3D point measured on simulator depth for the same pixel.

\paragraph{Implementation.}
Depth hypotheses lie on the query ray from 5\,mm below to 36\,cm above the table, 1.5\,mm apart.
At each one, the $15{\times}15$ gray-level window around the query pixel is warped into the second view through a horizontal plane or a plane facing the query camera, whichever matches better, and compared by normalized cross-correlation with Gaussian weights, so that the pixels nearest the query decide.
A match is accepted if its score $s$ is high, if it stands out from the best score elsewhere on the ray, $s_r$, by the ratio $\rho=(1-s_r)/(1-s)$, which rejects repeated structures, and if the window has texture (a gray-level spread of at least 4).
We require $s\ge0.85$ and $\rho\ge1.5$ for the current cameras and the previous observation, and $s\ge0.9$ and $\rho\ge2$ for older wrist frames (up to six observations back), since objects may have moved; a second view must be at least 2\,cm or $8^\circ$ away from the query camera.

\paragraph{Why not feature matching.}
Triangulating from feature matches between two views is a natural alternative, but it suits our queries poorly: the head and wrist cameras see the scene from very different directions and at very different scales, so local features often fail to match between them, and the pixel the agent queries, such as the center of a hole, need not carry a matchable feature at all, so its depth would have to be borrowed from matches nearby.
Our epipolar matching instead searches along the ray of the queried pixel itself and warps the window by the plane-induced homography, which absorbs the differences in viewpoint and scale.

\paragraph{Where it answers, it is accurate to millimeters.}
On all 1,182 query pixels of the Precision tasks (\cref{tab:sweep_offline}), epipolar matching answers 37\% of the pixels, using the other cameras and the wrist frames of the last six observations, at a median error of 4.6\,mm; two thirds of its answers lie within 1\,cm.
This is about four times more accurate than measuring object silhouettes (20.0\,mm) and ten times more than intersecting the pixel's ray with the table (51.6\,mm).
The gap is largest on holes, where a silhouette has no surface to measure: 2.8 \vs 36.4\,mm on \emph{deposit coin}.
Past wrist frames raise the coverage from 24\% to 37\% at no loss of accuracy, since a moving wrist camera provides new baselines for free.

\paragraph{Where it does not, the tool falls back and says so.}
Pixels on textureless, slanted or specular surfaces have no reliable match; there the tool reads the depth estimated by Depth Anything~3, accurate to about a centimeter, or the table plane, and tags each answer with its source.
The agent therefore gets an answer for every pixel and knows which ones to trust to millimeters: the full chain answers all 1,182 pixels, not only the 37\% that epipolar matching accepts, at a median error of 5.4\,mm, close to that of epipolar matching alone, and puts 64\% of them within 1\,cm, against 36\% for object silhouettes and 10\% for the table plane (\cref{tab:sweep_offline}).

\begin{table}[t]
  \centering
  \small
  \setlength{\tabcolsep}{3pt}
  \begin{tabular*}{\linewidth}{@{\extracolsep{\fill}}lccc@{}}
    \toprule
    Method & Coverage & Median & $\le$1\,cm \\
    \midrule
    Ray $\cap$ table plane & 100\% & 51.6\,mm & 10\% \\
    Object silhouettes & 97\% & 20.0\,mm & 36\% \\
    Epipolar, current views & 24\% & 4.8\,mm & 68\% \\
    Epipolar, + past wrists & 37\% & 4.6\,mm & 66\% \\
    \midrule
    \textbf{Ours} (epipolar, DA3, table) & \textbf{100\%} & 5.4\,mm & 64\% \\
    \bottomrule
  \end{tabular*}
  \caption{\textbf{Offline accuracy of 3D queries} on 1,182 query pixels of the Precision tasks, against the 3D point measured on simulator depth: share of pixels answered, median error and share within 1\,cm.}
  \label{tab:sweep_offline}
\end{table}

\subsection{Evolving notes over more rounds}
\label{sec:suppl_rounds}

In the experiment of \cref{fig:evolve}, the agent keeps evolving its notes over 20 additional rounds instead of merging and freezing them (\cref{sec:exp_analysis}), using layouts that differ from the evaluation layouts.
The resulting notes explain why performance stops improving and fluctuates: each task keeps at most ten notes, and once full, the agent mostly overwrites existing notes rather than adding new ones, replacing general lessons with accounts of the latest episode.
For example, a \emph{play stacking toy} note on grasping square pieces between occupied pegs ends up reading ``This success placed all three squares between a tall ring stack and a blue block\ldots''.
Merging keeps only lessons that several independent rollouts support, stated without layout details, and freezing stops the latest episode from rewriting them, which is why \method merges and freezes its notes.

\subsection{Failure analysis on RoboDojo}
\label{sec:suppl_failures}

\begin{figure*}[t]
  \centering
  \includegraphics[width=\textwidth]{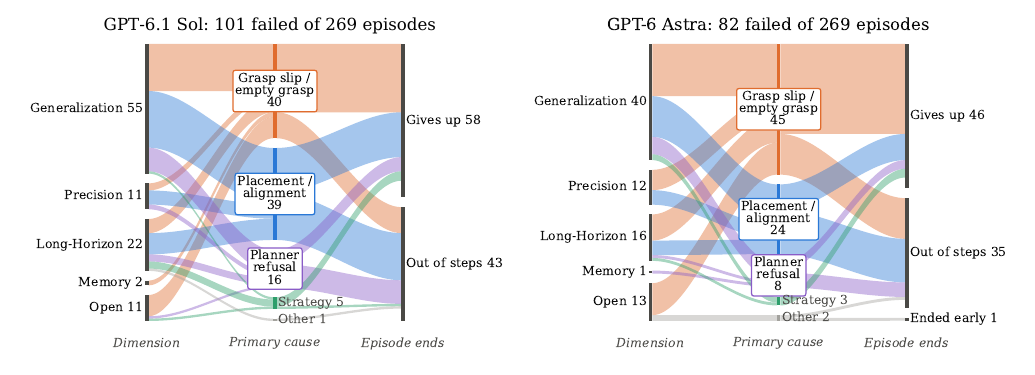}
  \caption{\textbf{Where RoboDojo failures come from.} Every failed episode of our runs with notes on layouts 0--4 of seed~0, including the randomized variants (269 episodes per model), flows from its dimension (left) to its primary cause (middle) and to how it ended (right); band widths are numbers of episodes, colored by cause. No failure is primarily attributed to a wrong 3D measurement or to memory.}
  \label{fig:failures}
\end{figure*}

We classify every failed episode of our runs with notes on layouts 0--4 of seed~0, standard and randomized variants (269 evaluated episodes per model), by its primary cause and by how it ended (\cref{fig:failures}).
The cause is labeled by an LLM from evidence extracted from the trace: grasp-monitor events, planner refusals, the sources of the queried points, and the reason and hindsight that the agent states when it gives up.

\paragraph{Contact, not perception.}
With either model, two causes account for more than three quarters of the failures: a grasp that slips or closes empty, and a placement or alignment that is off, such as a pen released on the rim of its holder or a keyboard pushed past its frame.
No failure is primarily due to a wrong 3D measurement, and none to memory: where the agent fails, it usually knows where things are and what remains to be done, but cannot hold or seat them.

\section{Prompts}
\label{sec:suppl_prompts}

\Cref{tab:prompts} summarizes the prompts of \method on RoboDojo, which are written in English.
The agent reads the first three at every call; the last four are used offline, once per rollout or per task.
On other robots only the robot description in the prompt changes (\cref{sec:exp_setup}).
Three principles recur across them.
\emph{Measure, do not guess:} every target is placed from \texttt{query\_points} on distinctive pixels, and each reading comes with its source.
\emph{Steps are scarce:} the prompts name the step budget, ask for final poses rather than small increments, and require every check or retry the agent or its notes add to be bounded.
\emph{General, not layout-specific:} guides tag what holds in any layout, and notes and rules state situations and actions without coordinates.

\begin{table*}[t]
  \centering
  \small
  \renewcommand{\arraystretch}{1.15}
  \begin{tabular}{@{}>{\raggedright\arraybackslash}p{0.13\textwidth}>{\raggedright\arraybackslash}p{0.13\textwidth}>{\raggedright\arraybackslash}p{0.69\textwidth}@{}}
    \toprule
    Prompt & Used & Main content \\
    \midrule
    System prompt & every call &
    The robot (two 6-DoF arms with parallel grippers, world frame, table height, the grasp point as the commanded position); the two budgets, 600 model calls and the environment's steps, and one tool call per turn; how to measure with \texttt{query\_points}: several distinctive pixels per object, on its top and on two opposite edges, and how far to trust each source; motion: name final poses, scale moves to the distance from contact, choose grasps and placement order with the later steps in mind, and move an idle wrist camera to see the work; the episode ends by itself on success, so the agent should not wait for it. \\
    Tool descriptions & every call &
    \texttt{move\_eef}: absolute targets naming only the dimensions to change, orientation from a straight-down grasp, gripper acting after arrival, an optional ramp (0--12 steps, longer to carry gently, shorter to strike), slip handling, up to five keyposes per call, and a one-sentence note on what the agent sees and why it acts. \texttt{query\_points}: up to 32 pixels of one camera, each answered with its source and accuracy, without using environment steps. \texttt{wait}: hold the arms for a number of steps and return keyframes. \texttt{give\_up}: end the episode with a reason and a hindsight note. \\
    First user turn & every call &
    The instruction, the official description of the task, the task guide with its keyframes, and the experience notes. \\
    \midrule
    Writing notes & after each writing rollout &
    Appended to the finished rollout with its outcome and the current notes. Record only general lessons that decided the outcome of the task or a sub-step; keep at most ten notes of at most 110 words, replacing or merging rather than adding; state when a lesson holds instead of unconditional rules; bound any verification or retry in attempts or steps; answer with one \texttt{update\_experience} call. \\
    Merging & once per task &
    From three note sets and the outcomes and summaries of their 15 rollouts, write at most five rules of at most 70 words, ``When \emph{situation}: \emph{action}'', each supported by at least two rollouts and cited, with a step cost where it spends steps; no unbounded verification, no layout-specific numbers, and any advice referring to unavailable tools is rewritten or dropped. \\
    General rules & once &
    From the rules of all 42 tasks, five to eight lessons (eight in our runs) that recur in at least five tasks, in the same form and at most 60 words each, without task or object names. \\
    Task guide & once per task &
    From one demonstration's trajectory and frames: a warning that the guide describes one demonstration in one layout; every statement tagged \textsc{general} or \textsc{demo layout}; an overview with the stages and the hard parts; and per stage its goal, the moves in \texttt{move\_eef} values, what was checked, and keyframes. \\
    \bottomrule
  \end{tabular}
  \caption{\textbf{Prompts of \method}, summarized. The first three form the agent's context at every call; the others write the task guides and the experience notes offline (\cref{sec:method:experience}).}
  \label{tab:prompts}
\end{table*}

\section{More experiments}
\label{sec:suppl_more}

\subsection{Controlled comparisons on a public subset}
\label{sec:suppl_subset}

To separate the factors behind \method's results, we rerun it with one factor changed at a time on a fixed subset: the 50 RoboDojo episodes whose step-wise GPT-6 Astra Direct runs are public~\cite{gptaspolicy}, ten tasks on seed~0 (41 standard and 9 randomized episodes).
One of them, layout~4 of \emph{imitate sorting sequence}, never starts because the simulator marks it unstable, so every configuration is evaluated on the same 49 episodes.
Unless stated otherwise, the configuration is that of the paper: GPT-6.1 Sol at effort \texttt{xhigh}, at most five keyposes per call, task guides, and the frozen merged rules, task and general, of every task.
All configurations of \cref{tab:subset} run in the same batch, including the default \method configuration, and are compared with it episode by episode (exact McNemar test).
The subset leans toward Long-Horizon (20 of 50 episodes) and has one task each from Precision, Memory and Open.
With 49 episodes, only large differences can be resolved: two configurations must differ by about ten episodes to be statistically distinguishable.

\begin{table*}[t]
  \centering
  \small
  \setlength{\tabcolsep}{4pt}
  \begin{tabular*}{\textwidth}{@{\extracolsep{\fill}}llcccccc@{}}
    \toprule
    & Configuration & Solved & SR (95\% CI) & Score & Calls & Cost & Paired ($p$) \\
    \midrule
    \multicolumn{2}{@{}l}{\method (guide, notes, $\le$5 keyposes, GPT-6.1 Sol \texttt{xhigh})} & 33/49 & 67.3 (53.4--78.8) & 77.9 & 37.8 & \$0.53 & -- \\
    \midrule
    \multirow{3}{*}{Inputs} & $-$ guide & 27/49 & 55.1 (41.3--68.1) & 67.8 & 39.5 & \$0.49 & 11\,/\,5 (0.21) \\
    & $-$ notes & 27/49 & 55.1 (41.3--68.1) & 67.1 & 37.3 & \$0.49 & 11\,/\,5 (0.21) \\
    & $-$ guide, $-$ notes & 19/49 & 38.8 (26.4--52.8) & 53.0 & 40.6 & \$0.46 & 20\,/\,6 (0.009) \\
    \midrule
    \multirow{2}{*}{Effort} & \texttt{high} & 27/49 & 55.1 (41.3--68.1) & 71.6 & 39.3 & \$0.49 & 12\,/\,6 (0.24) \\
    & \texttt{medium} & 30/49 & 61.2 (47.2--73.6) & 74.9 & 38.3 & \$0.41 & 8\,/\,5 (0.58) \\
    \midrule
    Model & Gemini 4 Argon (\texttt{high}) & 30/49 & 61.2 (47.2--73.6) & 70.9 & 29.4 & -- & 9\,/\,6 (0.61) \\
    \bottomrule
  \end{tabular*}
  \caption{\textbf{Controlled comparisons on the public RoboDojo subset} (49 episodes of ten tasks, seed~0), changing one factor of \method at a time. Solved episodes, success rate with Wilson 95\% interval, mean partial-credit score, model calls and cost (GPT-6.1 Sol list price) per episode, and the paired comparison with the first row: episodes solved only by it / only by this configuration (exact McNemar $p$). All rows run in the same batch. Gemini~4 Argon reads the notes written with GPT-6.1 Sol; its cost is omitted, as its price differs.}
  \label{tab:subset}
\end{table*}

\paragraph{Guides and notes.}
Removing either the guide or the notes costs six episodes (33 to 27), and removing both costs 14 ($p=0.009$).
The two inputs are thus partly interchangeable: without notes, the guide adds 8 episodes (19 to 27), and without the guide, the notes add 8 as well, while together they add 14.
Both convey how to approach the task, the guide from a demonstration and the notes from the agent's own rollouts, so either one recovers much of what the other provides.
Without either, \method still solves 19 of 49 episodes with GPT-6.1 Sol, compared with 13 of 50 for the step-wise GPT-6 Astra Direct with the stronger GPT-6 Astra.

\paragraph{Reasoning effort.}
Lowering GPT-6.1 Sol's effort from \texttt{xhigh} to \texttt{high} or \texttt{medium} lowers the cost per episode by 8\% and 22\%, while success changes only within the noise of 49 episodes.

\paragraph{Another model.}
With the same prompt, tools, guides and notes, Gemini~4 Argon (at \texttt{high}, the highest effort its endpoint accepts) solves 30 of 49 episodes, close to GPT-6.1 Sol and with fewer calls (29.4), so \method works with another frontier model without any change.

\subsection{Real robot}
\label{sec:suppl_real}

\paragraph{Setup.}
The setup follows \cref{sec:exp_setup}: a UR12e with a Robotiq 2F-85; a wrist Orbbec Gemini 335Lg and a fixed third-person Intel RealSense; the same agent, tools and executor as on RoboDojo; wrist stereo depth and cross-camera triangulation; and no pre-collected demonstrations, guides, notes or task-specific code.
For drawing, the cap $n_{\max}$ is raised from five to 30 keyposes per \texttt{move\_eef} call, which long strokes need; the portrait instruction states this cap explicitly.
Each task is given as one natural-language instruction.
When a task involves a person, the agent lets the person act with \texttt{wait} and sees the frames recorded meanwhile.
Success is verified by a person against a fixed criterion per task, such as a stable four-level stack, a fully seated workpiece or five marks in a row, and confirmed on the final camera frames; 75 of the 85 episodes succeed (\cref{tab:real}).
Whether a task ran with GPT-6.1 Sol or GPT-6 Astra was assigned randomly before the experiment and kept fixed; the prompt was also finalized before any real-robot run and was never tuned on these tasks.
We also tried the GPT-6 Astra (RoboProbe) baseline~\cite{astra} once per task; a single episode usually took more than 30 minutes at a high API cost, and none succeeded, so we did not scale it up or report it in \cref{tab:real}.
Every episode is judged for success; per-episode statistics (model calls, keyposes, queried pixels and time) in \cref{tab:real} and \cref{tab:real_full} are computed over the episodes with full logs, one or more per task.
\Cref{tab:real_full} details each of the 16 tasks, and \cref{sec:suppl_vis} shows one successful episode per task.

\paragraph{3D queries on the real robot.}
A queried pixel takes the FoundationStereo~\cite{wen2025foundationstereo} depth of the wrist camera's infrared pair around it, then cross-camera triangulation between the wrist and third-person cameras where that depth is missing, and falls back to the table plane, each answer tagged with its source and warned when noisy.
The wrist camera is calibrated by hand-eye calibration (Tsai--Lenz) on a chessboard, and the third-person camera's pose is calibrated from chessboards viewed jointly by both cameras.

\begin{table*}[t]
  \centering
  \small
  \setlength{\tabcolsep}{3pt}
  \begin{tabular*}{\textwidth}{@{\extracolsep{\fill}}llrrrrrrc@{}}
    \toprule
    Task & Family & Calls & \texttt{move\_eef} & Keyposes & KP / move & Pixels & Min & Success \\
    \midrule
    Four-level stack & Long-Horizon & 28 & 10 & 23  & 2.3  & 83  & 5.2  & 5/5 \\
    Nest cups, drop in cap            & Long-Horizon & 22 & 8  & 19  & 2.4  & 90  & 7.2  & 5/5 \\
    Two apples onto a plate                     & Long-Horizon & 11 & 5  & 13  & 2.6  & 37  & 1.9  & 5/5 \\
    Peg-in-hole insertion                       & Precision    & 21.3 & 7.7 & 12  & 1.6  & 67  & 5.5  & 2/5 \\
    Pour cat food onto a plate                  & Precision    & 13 & 8  & 21  & 2.6  & 26  & 3.7  & 5/5 \\
    Shell game                                  & Memory       & 15 & 6  & 12  & 2.0  & 26  & 3.8  & 5/5 \\
    Draw the NVIDIA Voyager                     & Open         & 26 & 16 & 65  & 4.1  & 34  & 7.6  & 5/5 \\
    Draw a house                                & Open         & 12 & 6  & 45  & 7.5  & 31  & 4.1  & 5/5 \\
    Copy a rocket in 3 colors       & Open         & 29 & 15 & 125 & 8.3  & 90  & 8.5  & 4/5 \\
    Draw a portrait                             & Open         & 23 & 15 & 305 & 20.3 & 24  & 10.1 & 3/5 \\
    Free-form building      & Open         & 30 & 17 & 58  & 3.4  & 92  & 9.0  & 5/5 \\
    Cup pyramid in context       & In-context   & 20 & 12 & 40  & 3.3  & 59  & 7.1  & 5/5 \\
    Block castle in context          & In-context   & 30 & 12 & 40  & 3.3  & 110 & 10.2 & 5/5 \\
    Gomoku endgame \vs a person                 & Gomoku \vs human & 21 & 8  & 51  & 6.4  & 48  & 7.8  & 5/5 \\
    Full Gomoku game \vs a person               & Gomoku \vs human & 33 & 12 & 98  & 8.2  & 61  & 12.4 & 5/5 \\
    Fold a shirt                                & Shirt folding & 17 & 10 & 26  & 2.6  & 32  & 3.5  & 6/10 \\
    \midrule
    Logged episodes (mean) & & 21.9 & 10.4 & 57.3 & 5.5 & 56.6 & 6.6 & \textbf{75/85} \\
    \bottomrule
  \end{tabular*}
  \caption{\textbf{Real-robot tasks}, one row per task. Calls: model calls; \texttt{move\_eef}: keypose sequences; KP / move: keyposes per sequence; Pixels: pixels queried; Min: minutes. These columns are computed over the logged episodes of each task. Success: successful episodes of the task over all its episodes, five per task and ten for the shirt. Last row: means over all logged episodes, computed as in \cref{tab:real} (KP / move: total keyposes over total sequences), and total successes over all 85 episodes.}
  \label{tab:real_full}
\end{table*}

\paragraph{The agent sets the keypose density.}
Keyposes per \texttt{move\_eef} call range from 1.6 for insertion, where the agent moves in short sequences and looks after each, to 20.3 for the portrait (\cref{tab:real_full}).
A grasp and place takes three or four keyposes, sent in sequences of 2.3--2.6 on the Long-Horizon tasks, close to RoboDojo, while drawing grows with the length of the strokes: 45 keyposes for a house, 125 for the rocket and 305 for the portrait.
Model calls vary far less, from 11 to 33 per episode, so the cost of an episode is set by how often the agent needs to look, not by how much it draws.

\paragraph{Failures.}
The ten failures fall on shirt folding (4 of 10), precision (3 of 10) and open tasks (3 of 25).
Most concern contact: an insertion releases before the pegs are seated, or a shirt slides along with the gripper during a one-arm fold.
The others concern the clearance of the robot's own body: lowering a pen grasped near its tip would have brought the wrist camera into collision with the table, and the robot stopped before contact; this is also why the portrait and Gomoku instructions, written before the experiment like all others, ask to grasp the pen near its end.
As on RoboDojo, these call for force or tactile feedback at contact and a collision check of the whole arm and camera before a sequence is sent.

\paragraph{Time.}
A logged episode takes 6.6 minutes on average; a model call takes a median of 11.7\,s and executing its keypose sequence 6.1\,s, so inference, not motion, sets the pace, which confines \method to quasi-static scenes (\cref{sec:conclusion}).

\subsubsection{The agent's reasoning}
\label{sec:suppl_reasoning}

Every tool call carries the agent's note on what it sees and why it acts (\cref{sec:suppl_impl}).
We quote notes where the agent does more than the instruction spells out.

\newcommand{\agentquote}[1]{\begin{quote}\small\itshape #1\end{quote}}

\paragraph{Setting up a double threat.}
In the Gomoku endgame, before its first move, the agent plans two lines at once:
\agentquote{``Black has the diagonal three C3, D4, E5 as well as F4 and F5; I choose F6 to make a diagonal four and a vertical threat.''}
The person blocks the diagonal at G7, and the agent turns to the second line:
\agentquote{``White's new move at G7 confirmed; black draws an x on the empty F3, so that F3--F6 form a four open at both ends.''}
An open four cannot be blocked at both ends; the person blocks F7 and the agent wins at F2.
In the full game it adapts in the same way: ``White blocked the three on the diagonal at its lower-right end. I switch to the horizontal line'', and at the end it races the person: ``White's latest move forms a four above, still not five. It is black's turn: I draw an x at the measured, still empty right end, completing five horizontally.''

\paragraph{Noticing an incomplete demonstration.}
In the castle task, the person first built the castle but did not yet scatter the blocks. Rather than start rebuilding, the agent keeps watching:
\agentquote{``The first demonstration showed the finished castle but lacked the scattering step that the task requires; keep the arm still and watch this part as well.''}

\paragraph{Trusting the images over the gripper.}
Folding the shirt, the gripper reports an empty grasp on the thin hem. The agent does not retry blindly:
\agentquote{``The thin hem may be held even though the gripper reports nothing. I will make a small vertical lift and check whether the fabric follows before retrying.''\\
``The images confirm that the hem lifted despite the empty-grasp report. I will carry the lower hem in an arc over the shirt's midpoint to fold the lower half upward.''}

\paragraph{Closing the loop through the ink.}
In every drawing task the agent first draws a short test stroke, looks at it, and adjusts the pen height by millimeters:
\agentquote{``The top view shows the lower half of the C touched too lightly and the line is broken. Lower the pen by only 1.5\,mm and complete these three straight strokes, without adding lateral pressure.''}
Copying the rocket, it measures each marker separately after each regrasp: ``The red nib is about 2.4\,mm shorter below the grasp than the black nib. I will draw the two straight red nose edges at the corresponding contact height'', and later ``The blue nib is another 1.5\,mm shorter at this grasp.''
For the portrait it respects the cap of the instruction exactly, noting that its last drawing call ``has exactly 30 keyposes''.

\paragraph{Tracking under occlusion.}
In the shell game the agent states its goal before the person starts, ``observe which cup the cap is hidden under, and continuously track that cup's position during the shuffle'', and confirms it on lifting the first cup it chose: ``The red cap has appeared from under this cup, confirming that the tracking was correct.''

\paragraph{Diagnosing its own failure.}
In insertion trial 1 the agent lowered the workpiece 4\,mm into the holes, released it and lifted; the arm then entered a protective stop. It measures the result, gives up rather than waiting for a correction it cannot make, and states why and what it should have done:
\agentquote{``The pin workpiece is upright on the receiving workpiece and the gripper is open and clear, but the final height measurements indicate that the pegs are not fully seated. The robot is now in protective-stop safety mode, so I cannot safely correct the insertion.''\\
Hindsight: ``The insertion should have been checked for actual downward movement of the workpiece at the first contact step. A better alignment check and verified peg engagement were needed before releasing.''}
In the two successful trials it stops at first contact and lets gravity seat the workpiece:
\agentquote{``The pin is now supported at the hole block, and its small movement relative to the jaws suggests contact. I will not push farther: I will release it to settle naturally.''\\
``I will lower another 7\,mm to establish roughly 1\,cm of engagement, then release rather than press the shoulder against the receiver.''}
This is the kind of lesson that the experience notes of \cref{sec:method:experience} distill on RoboDojo.

\subsection{RoboTwin 2.0}
\label{sec:suppl_robotwin}

RoboTwin 2.0~\cite{chen2025robotwin2} has 50 bimanual tasks on the Aloha-AgileX robot and a public leaderboard.
Its entries are trained on 50 demonstrations per task in clean scenes and tested in two settings: Easy, the same clean scenes, and Hard, randomized scenes with distractor objects, new textures and lighting, and a lower table.
Hard thus asks whether a policy still works when the scene looks different from its training data.

\paragraph{Setup.}
We run \method with the same agent, tools and executor and with GPT-6.1 Sol, using only the RGB images of the head and wrist cameras, ten episodes per task and setting (1,000 in all).
As on RoboDojo, it reads a guide written from one demonstration of each task; unlike on RoboDojo, its notes are not merged in advance but written after every episode for the later episodes of the same task.
Two small additions handle the platform: the prompt says that a grasp shows only in the images, since RoboTwin's gripper reading always reads closed after a grasp command, and the table height, which varies between scenes, is measured once at the start by triangulating table points between the cameras.

\begin{table}[t]
\centering
\small
\setlength{\tabcolsep}{2pt}
\begin{tabular*}{\linewidth}{@{\extracolsep{\fill}}llcccc@{}}
\toprule
Method & Demos & Easy & Hard & $\Delta$ & \textbf{Avg.} \\
\midrule
ACT~\cite{zhao2023act} & 50/task & 29.7 & 1.7 & $-28.0$ & 15.7 \\
DP~\cite{chi2023diffusionpolicy} & 50/task & 28.0 & 0.6 & $-27.4$ & 14.3 \\
DP3$^\ddagger$~\cite{ze2024dp3} & 50/task & 55.2 & 5.0 & $-50.3$ & 30.1 \\
RDT~\cite{liu2024rdt} & 50/task & 34.5 & 13.7 & $-20.8$ & 24.1 \\
$\pi_0$~\cite{black2024pi0} & 50/task & 46.4 & 16.3 & $-30.1$ & 31.4 \\
\midrule
$\pi_{0.5}$~\cite{black2025pi05} & 50/task$^\dagger$ & 70.7 & 46.0 & $-24.7$ & 58.3 \\
PrismWAM$^\ddagger$ & 50/task$^\dagger$ & 91.8 & 75.5 & $-16.3$ & 83.7 \\
GO-3 Nano Preview & 50/task$^\dagger$ & 88.5 & 81.5 & $-7.1$ & 85.0 \\
Awomo-0.6 & 50/task$^\dagger$ & 88.2 & \underline{84.8} & $-3.4$ & 86.5 \\
CloudRobo-WM & 50/task$^\dagger$ & \textbf{95.3} & 82.7 & $-12.6$ & \underline{89.0} \\
\midrule
\method (ours) & 1 demo & \underline{94.6} & \textbf{95.0} & $+0.4$ & \textbf{94.8} \\
\bottomrule
\end{tabular*}
\caption{\textbf{RoboTwin 2.0}, success rate (\%) over 50 tasks in clean (Easy) and randomized (Hard) scenes. Leaderboard rows: snapshot 2026-11-15, 100 trials per task, 50 demonstrations per task ($^\dagger$all tasks jointly; $^\ddagger$with depth). \method: ten episodes per task, GPT-6.1 Sol, one demonstration per task as a guide. \textbf{Bold}: best; \underline{underline}: second.}
\label{tab:robotwin}
\end{table}

\paragraph{Results.}
\method succeeds in 94.6\% of the episodes in Easy and 95.0\% in Hard (\cref{tab:robotwin}).
In Easy it is on par with the best entry (95.3\%); in Hard it leads the best entry by 10.2 points.
Randomization does not hurt it at all ($+0.4$ points), whereas every leaderboard entry in \cref{tab:robotwin} loses from Easy to Hard, by 3.4 to 50.3 points.
The reason is how each acts: a learned policy maps pixels to actions and has seen only clean scenes, so clutter and new textures shift its input, whereas \method names the objects it needs, measures them and plans from the measurements, so a distractor only changes which pixels it asks about.

\paragraph{Failures and cost.}
Almost all of the 52 failures are again about contact: an object slips, tips over or is pushed away, often out of view, and the agent gives up after searching for it; they concentrate on \emph{place dual shoes}, where placing the second shoe often knocks over the first.
An episode takes 8 model calls at the median and costs about \$0.17.
Within ten episodes per task and so close to the ceiling, the notes give no measurable gain.

\paragraph{Caveats.}
Ten episodes per task, instead of the leaderboard's 100, leave about one point of uncertainty on the overall rate but much more on a single task.
\method reads one demonstration per task instead of training on 50 and uses an API-served model, so the comparison is between approaches rather than between policies trained on the same data.

\section{Visualizations}
\label{sec:suppl_vis}

\begin{figure*}[p]
  \centering
  \includegraphics[width=\textwidth]{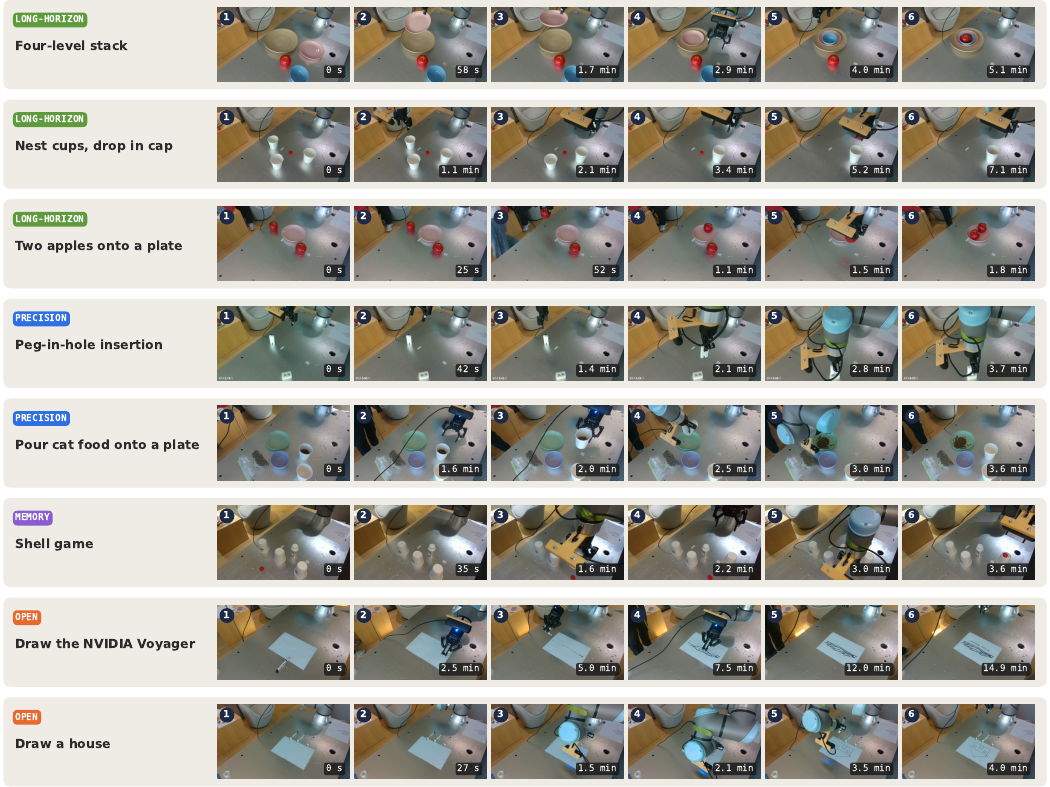}
  \caption{\textbf{Real-robot episodes (1/2):} Long-Horizon, Precision, Memory and the first Open tasks. Six third-person keyframes per episode, with the time since its start.}
  \label{fig:real_rows_1}
\end{figure*}

\begin{figure*}[p]
  \centering
  \includegraphics[width=\textwidth]{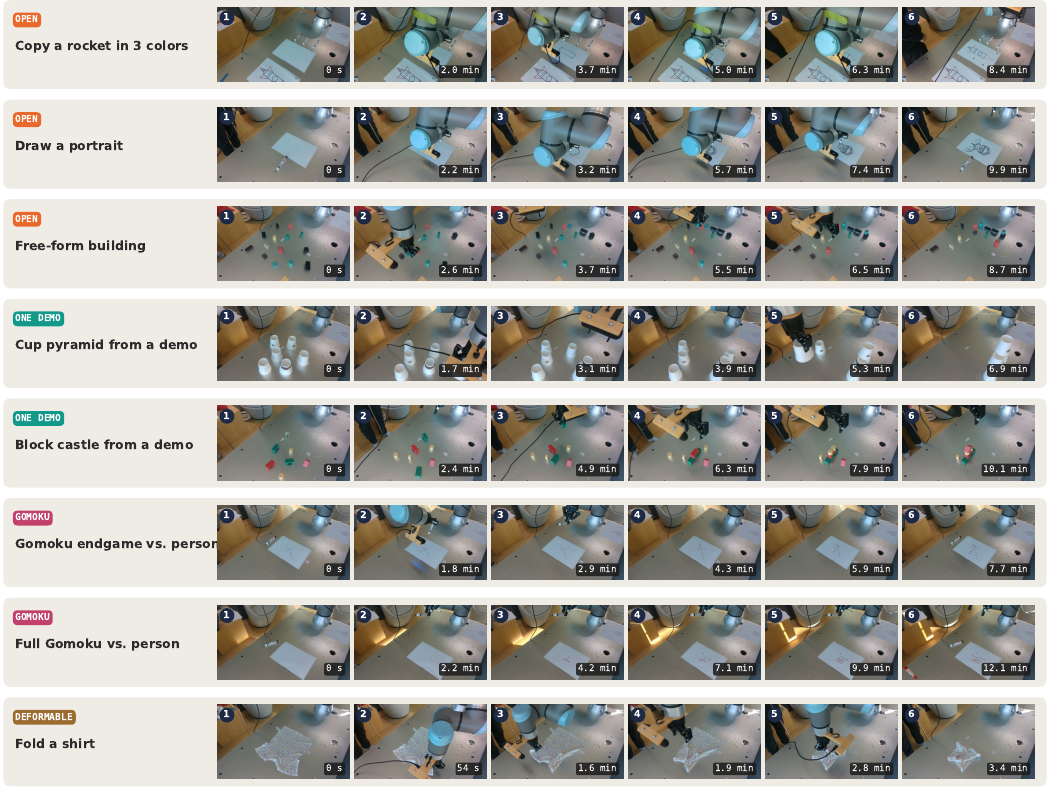}
  \caption{\textbf{Real-robot episodes (2/2):} the remaining Open tasks, in-context rebuilding, Gomoku against a person and shirt folding.}
  \label{fig:real_rows_2}
\end{figure*}

\Cref{fig:real_rows_1,fig:real_rows_2} show one successful episode for each of the 16 real-robot tasks as six third-person keyframes timestamped from the start of the episode.

\begin{figure*}[p]
  \centering
  \includegraphics[width=\textwidth]{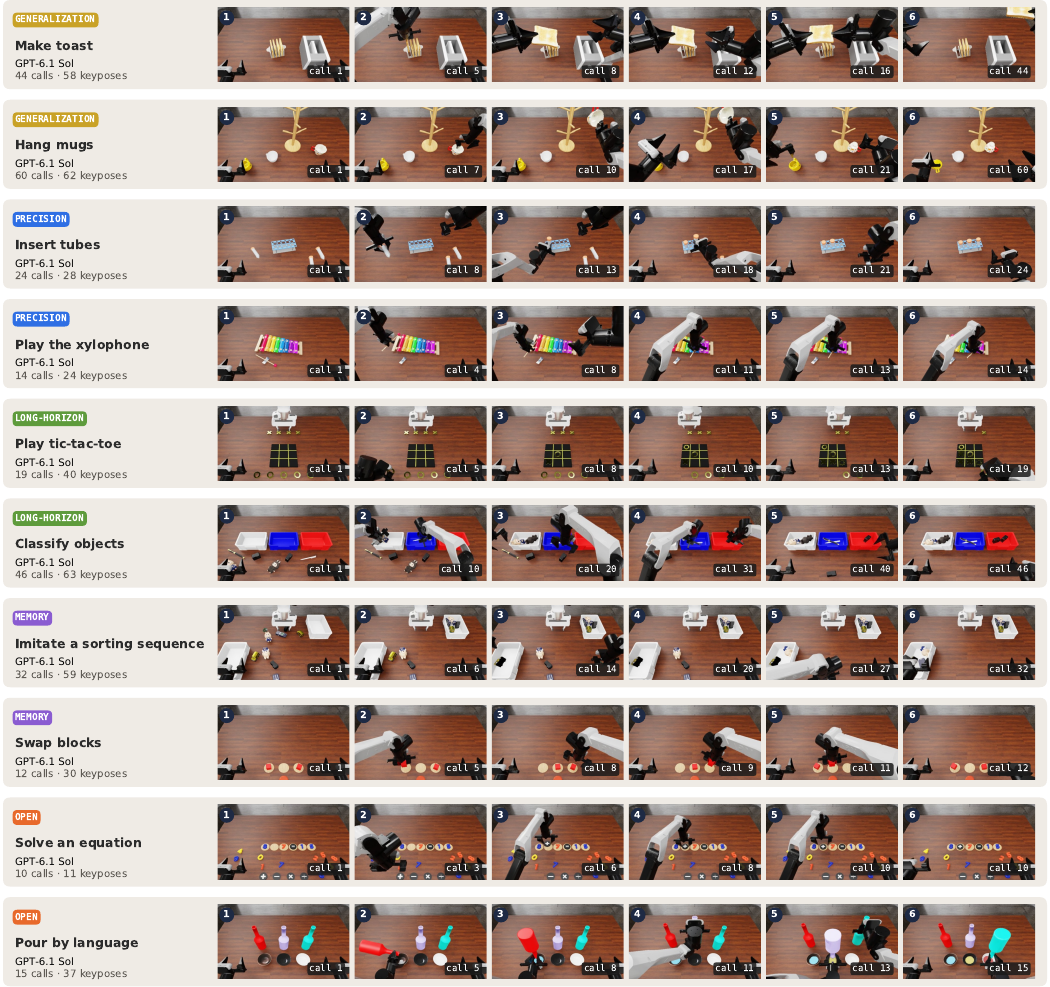}
  \caption{\textbf{RoboDojo episodes:} two tasks of each dimension, solved by \method with GPT-6.1 Sol. Six head-camera keyframes per episode, each with the model call that sees it.}
  \label{fig:sim_rows_robodojo}
\end{figure*}

\begin{figure*}[p]
  \centering
  \includegraphics[width=\textwidth]{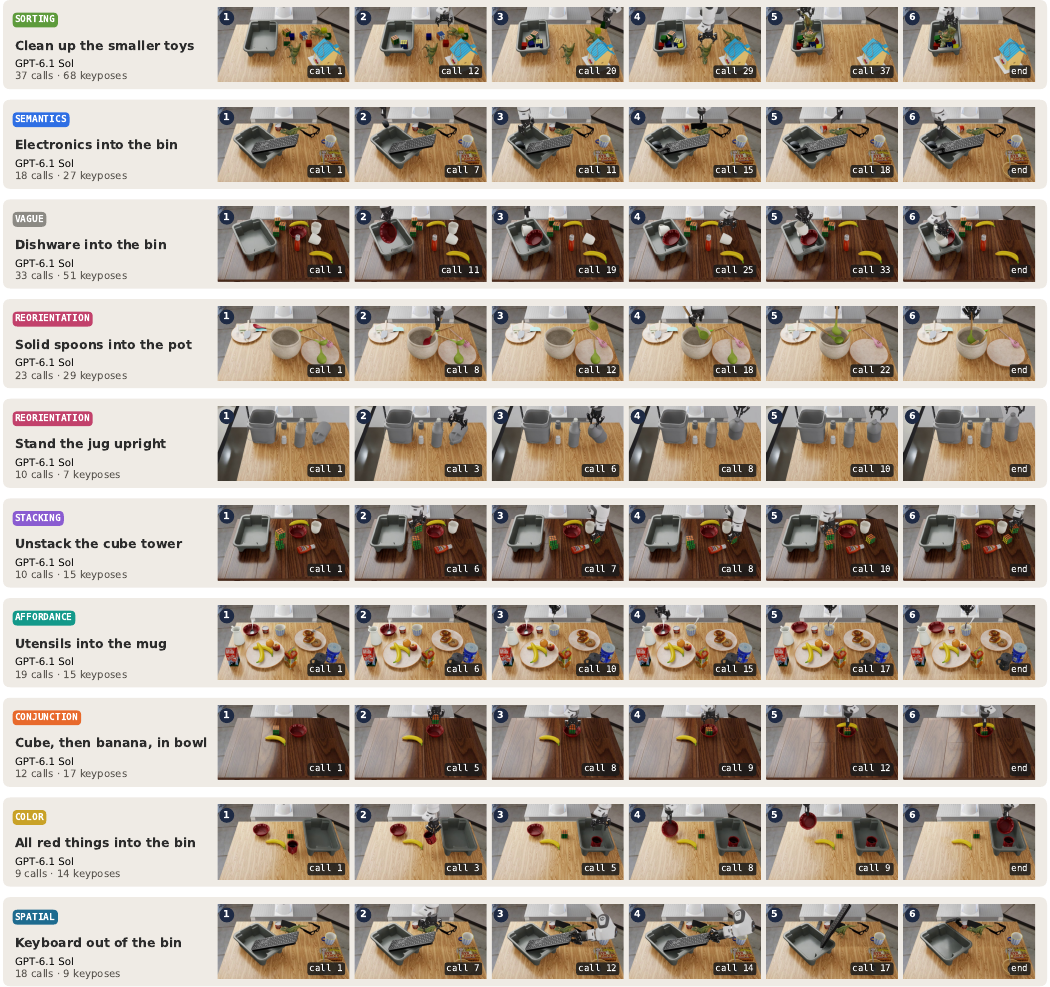}
  \caption{\textbf{RoboLab episodes:} ten tasks, labeled with one of their RoboLab attributes, solved by \method with GPT-6.1 Sol. Six viewport keyframes per episode, at the model calls shown and at the end of the episode.}
  \label{fig:sim_rows_robolab}
\end{figure*}

\begin{figure*}[p]
  \centering
  \includegraphics[width=\textwidth]{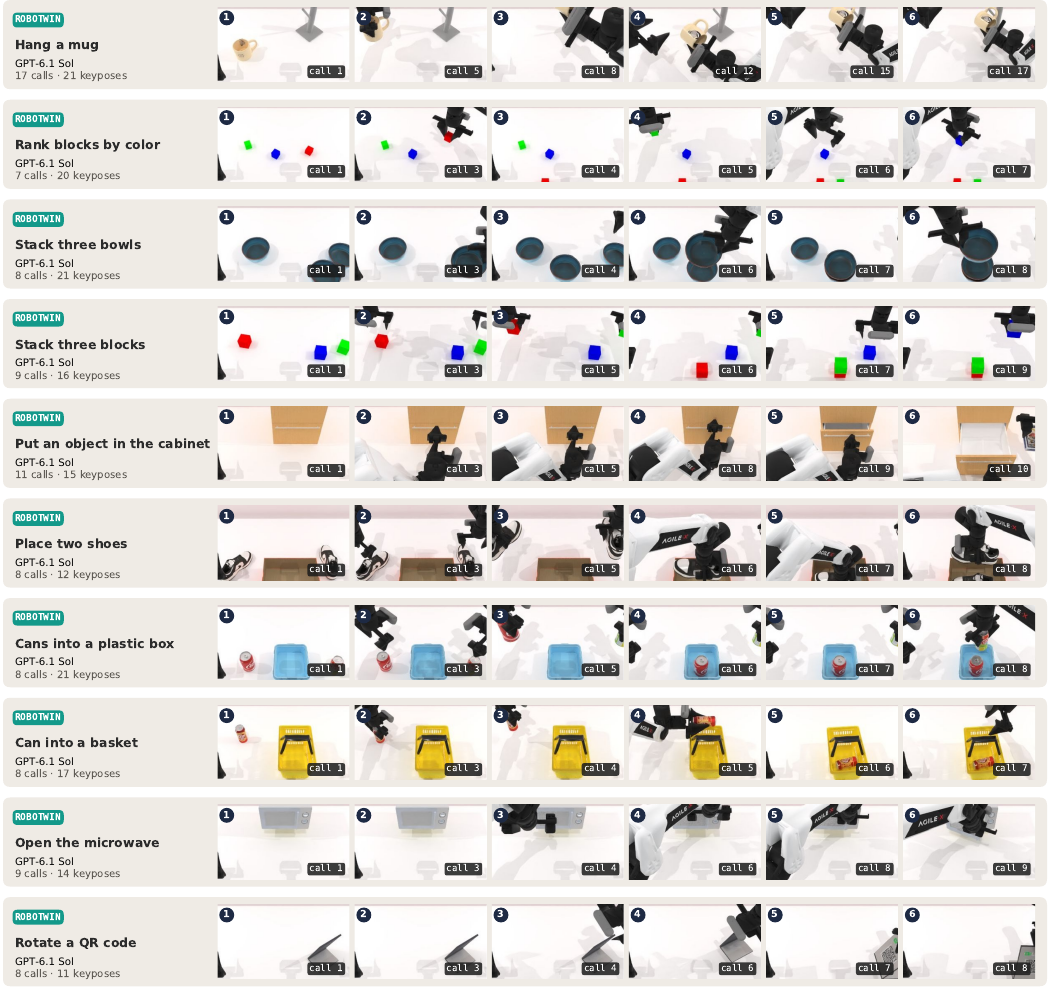}
  \caption{\textbf{RoboTwin~2.0 episodes} in the Easy setting, solved by \method with GPT-6.1 Sol. Six head-camera keyframes per episode, each showing the observation seen at the indicated model call, so the last frame precedes the final motion.}
  \label{fig:sim_rows_robotwin}
\end{figure*}

\Cref{fig:sim_rows_robodojo,fig:sim_rows_robolab,fig:sim_rows_robotwin} show ten solved tasks of each simulated benchmark, with GPT-6.1 Sol.
On RoboDojo we take two tasks of each dimension from our runs with notes, on RoboLab tasks with different attributes from the run with self-evolved notes behind \cref{tab:robolab}, and on RoboTwin~2.0 tasks of the Easy setting (\cref{sec:suppl_robotwin}); for each task we show the solved episode with the fewest model calls.
Each keyframe is labeled with the model call that sees it; on RoboLab, the last one shows the end of the episode.

\end{document}